\documentclass[runningheads]{llncs}

\usepackage{eccv}

\usepackage{eccvabbrv}

\usepackage{graphicx}
\usepackage{booktabs}
\usepackage{array}
\usepackage{booktabs} % For formal tables
\usepackage{bm}
\usepackage{float}
\usepackage{times,amsmath,epsfig}
\usepackage{graphicx}%插入图片
\usepackage[dvipsnames]{xcolor}
\usepackage{multirow} %多行合并
\usepackage{array} %\调用公式宏包的命令应放在调用定理宏包命令之前，也能控制表格
\usepackage{booktabs} %调整表格线与上下内容的间隔
\usepackage{longtable}%调用跨页表格
\usepackage{bm}%数学字体加粗
\usepackage{verbatim} %多行注释
\usepackage{float}

\usepackage{color}
\newcolumntype{H}{>{\setbox0=\hbox\bgroup}c<{\egroup}@{}}
\usepackage{colortbl}
\usepackage{pifont}
\newcommand{\cmark}{$\checkmark$}%
\newcommand{\xmark}{ }%
\usepackage{makecell}
\newcommand{\greenuparrow} {\textbf{\textcolor{ForestGreen}{ $\bm{\uparrow}$}}}
\newcommand{\greendownarrow} {\textbf{\textcolor{ForestGreen}{ $\bm{\downarrow$}}}}

\usepackage{arydshln}
\usepackage[accsupp]{axessibility}  % Improves PDF readability for those with disabilities.
\usepackage{bbding}

\usepackage{hyperref}

\usepackage{orcidlink}

\makeatletter
\def\thanks#1{\protected@xdef\@thanks{\@thanks
		\protect\footnotetext{#1}}}
\makeatother
\begin{document}

\title{Free-VSC: Free Semantics from Visual Foundation Models for Unsupervised Video Semantic Compression} 

\titlerunning{Free-VSC}

\author{Yuan Tian\inst{1,2}\orcidlink{0000-0001-6073-8582}   \and
Guo Lu\inst{1}\orcidlink{0000-0001-6951-0090}\textsuperscript{\Envelope} \and
Guangtao Zhai\inst{1}\orcidlink{0000-0001-8165-9322}\textsuperscript{\Envelope}   \thanks{\textsuperscript{\Envelope} Corresponding authors}}

\authorrunning{Tian et al.}

\institute{
 Institute of Image Communication and Network Engineering, Shanghai Jiao Tong University 
 \and 
 Shanghai Artificial Intelligence Laboratory \\
  \email{tianyuan168326@outlook.com, \{ luguo2014, zhaiguangtao\}@sjtu.edu.cn}
}
 
%\\

\maketitle

\begin{abstract}
  Unsupervised video semantic compression (UVSC), i.e., compressing videos to better support various analysis tasks, has recently garnered attention. However, the semantic richness of previous methods remains limited, due to the single semantic learning objective, limited training data, etc.
To address this, 
we propose to boost the UVSC task by absorbing the off-the-shelf rich semantics from VFMs.
Specifically, we introduce a VFMs-shared semantic alignment layer, complemented by VFM-specific prompts, to flexibly align semantics between the compressed video and various VFMs. This allows different VFMs to collaboratively build a mutually-enhanced semantic space, guiding the learning of the compression model.
Moreover, we introduce a dynamic trajectory-based inter-frame compression scheme, which first estimates the semantic trajectory based on the historical content, and then traverses along the trajectory to predict the future semantics as the coding context. 
This reduces the overall bitcost of the system, further improving the compression efficiency.
Our approach outperforms previous coding methods on three mainstream tasks and six datasets.
\end{abstract}

%\vspace{-8mm}
\section{Introduction}
\label{sec:intro}

Video compression methods including traditional~\cite{sullivan2012overview,bross2021overview} and learnable~\cite{lu2019dvc,hu2021fvc,li2021deep,li2023neural} ones, have achieved remarkable advances in terms of visual quality metrics such as PSNR.
However, directly deploying these methods to downstream analysis tasks usually achieves undesirable performance~\cite{yi2021benchmarking}, due to not particularly preserving semantics during compression, which calls for efforts on video semantic compression.

\begin{figure}[!t]
	\scriptsize
		\renewcommand\arraystretch{1.2}
	\tabcolsep=0.5mm
	\centering
		\begin{tabular}{cccc}  
			\includegraphics[width=0.225 \textwidth]{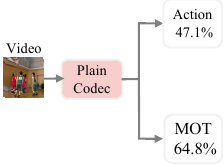}&
			\includegraphics[width=0.248 \textwidth]{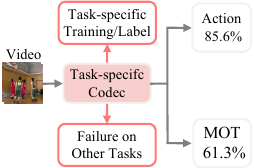}&
			\includegraphics[width=0.245 \textwidth]{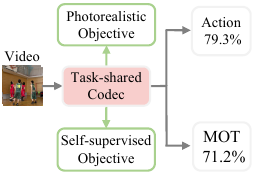}&
			\includegraphics[width=0.240 \textwidth]{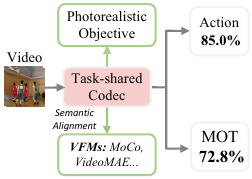}\\
			(a) Plain codec & (b) Supervised (unpractical)&(c) SSL-driven& (d) VFMs-driven (Ours)  \\ 
%			\includegraphics[width=0.42\textwidth]{fig/paradigm3.pdf}&
%			\includegraphics[width=0.42\textwidth]{fig/paradigm4.pdf}\\
%			\centering(c) SSL-driven & (d) VFMs-driven (Ours)
		\end{tabular}
	\vspace{-2mm}
	\caption {Recent video semantic compression paradigms, where (a) deploying plain codecs like VVC directly to downstream tasks yields inferior results, (b) Adapting learnable codecs with task-specific supervisions, such as action recognition, requires pre-deployment training and data labels, yet performs poorly on other tasks like multiple object tracking (MOT), thus unpractical, (c) prior unsupervised methods driven by self-supervised learning (SSL) exhibit undesirable results due to the limited learned semantics, and (d) our approach, driven by pre-trained visual foundation models (VFMs), achieves strong results by absorbing their rich semantics. Evaluation is conducted on UCF101@0.02bpp and MOT17@0.01bpp, utilizing TSM~\cite{lin2019tsm} and ByteTrack~\cite{zhang2021bytetrack}, respectively. Network architectures of (b)/(c)/(d) are consistent for a fair comparison.
	}
	\vspace{-7mm}
	\label{fig:review_paradigm}
\end{figure}

The previous works can be roughly divided into {supervised} and {unsupervised} ones.
{Supervised} methods, as shown in Figure~\ref{fig:review_paradigm} (b),
are typically coupled with target tasks, optimizing the codec with task models~\cite{choi2022scalable11,huang2022hmfvc} together.
The codec has to be adapted for the new task by a time-consuming supervised-training procedure, before being deployed.
More seriously, this usually leads to the performance degradation on other tasks.
The problems above limit their practicality.
Although there are some image-oriented methods~\cite{feng2022image,chen2023transtic} that enable a multiple tasks-shared encoder, their decoder part shall still be tuned towards the target task before deployment.
The tuning procedure is non-trivial, considering that
video annotations are laborious, and jointly fine-tuning codecs with video analysis models consumes heavy computational cost.

Therefore, recently, there emerges the {unsupervised} video semantic compression paradigm~\cite{tian2023non}, as shown in Figure~\ref{fig:review_paradigm} (c), aiming to produce low-bitrate videos that 
%are
%This paradigm dispenses with the need for task-specific guidance during compression, while 
%capable of 
can support various video analysis tasks.
% without task-specific training procedure. 
This paradigm optimizes a trade-off objective between the bitcost and the semantic distortion of the compressed video, along with a photorealistic term for regularizing the video visual quality.
The primary challenge lies in representing the semantic content of videos, so that a proxy function of semantic distortion can be quantitatively calculated.
Some early methods representing video semantics through hand-crafted representations, such as region-of-interest (ROI)~\cite{galteri2018video,cai2021novel} or segmentation maps~\cite{akbari2019dsslic,duan2022jpd},
unable to enjoy the benefits of big data.
Recently, a few approaches~\cite{fang2022prior,tian2023non} have integrated self-supervised learning objectives into compression. Although data-driven, these methods still face limitations in learning rich semantics, due to the adoption of simple and single learning objectives, limited training data, etc.

On the other hand, there has been a rapid advancement in the community of open-sourced Visual Foundation Models (VFMs)~\cite{he2020momentum,he2022masked,tong2022videomae,wang2023videomae}. These substantial-parameter models are trained with diverse learning objectives and large-scale datasets, probably emerging strong and complementary semantic representation capabilities~\cite{caron2021emerging}. This prompts a question: \textit{can we harness the rich semantic representation within VFMs for the unsupervised video semantic compression problem?}

Our answer is YES. We propose a new semantic coding framework termed Free-VSC to absorb free and rich semantics from multiple VFMs.
However, it is non-trivial to perform learning from several large-scale foundation models, considering the heterogeneity among their learned semantics.
To cope with this, we introduce a prompt-based semantic alignment layer (Prom-SAL) to align the compressed video semantics to a shared representation space of various foundation models, as shown in Figure~\ref{fig:review_paradigm} (d).
The parameters of Prom-SAL are shared for all VFMs, implicitly building a mutually-enhanced and harmonious semantic space, while the input prompts to Prom-SAL are VFM-specific and modulate the alignment procedure to adapt to different VFMs.

As for the compression of inter-frame semantic redundancies, most previous compression methods~\cite{cai2021novel,tian2023non,huang2022hmfvc} adopt the motion-based schemes.
However, 
unlike low-level pixels/features that show obvious motions (spatial displacements),
the temporal patterns of high-level semantic features are implicit and highly-related to the video content.
Thus, we propose to predict the next frame by first predicting the video-adaptive semantic trajectory, instead of relying on explicit motion.
The trajectories are modeled as distributions of non-linear transformations, which are dynamically predicted based on the historical information.
Then, we transform the previous features along the trajectory to obtain a preliminary future prediction, substantially reducing the uncertainty of the future frame, as well as its bitcost~\cite{shannon1948mathematical}.
Compared to previous motion or static transformation-based ones, the proposed scheme is more effective for semantic compression, due to its content-adaptiveness and flexible handling of high-level features.
Our contributions are:

\noindent \textbf{1.} We propose a new video semantic compression framework by harnessing powerful Visual Foundation Models (VFMs), reusing their rich and free semantic representations.
This is the first work employing VFMs for semantic compression. 

\noindent \textbf{2.} A prompt-based semantic alignment layer is devised to facilitate the framework learning mutually-enhanced semantics from multiple VFMs, effectively guiding the compression model to particularly preserve the semantics within videos.

\noindent \textbf{3.} A dynamic trajectory-based entropy model is introduced to remove inter-frame semantic redundancy, by first predicting video-adaptive trajectories within semantic space,
providing better semantic compression efficiency than other widely-used schemes.

\noindent \textbf{4.} Our approach demonstrates substantial improvements over previous methods on three tasks and six datasets.

\vspace{-4mm}
\section{Related Works}
\vspace{-4mm}
\textbf{Image/Video Semantic Compression.}
Traditional~\cite{wiegand2003overview,sullivan2012overview,bross2021overview,zhang2011parametric}, learnable codecs~\cite{balle2018variational,zhang2023lvqac,zhang2024gaussianimage,hu2021fvc,lu2019dvc,lu2020end,li2021deep,li2023neural}, and some tradition-neural mixed codecs~\cite{tian2021self,tian2023clsa} are designed for achieving good visual quality, instead of preserving video semantics.
% have achieved remarkable advances. 
%However, the above methods have not achieved satisfactory results on image/video-based analysis tasks.
To address this, early video coding for machine (VCM) standards such as CDVA~\cite{duan2013compact} and CDVS~\cite{duan2015overview,duan2018compact} transmit image keypoints, serving image indexing and retrieval tasks.
Later, several methods~\cite{zhang2016joint,choi2018near,chen2019lossy,chen2019toward,singh2020end,feng2022image,lin2023deepsvc} propose to compress intermediate feature maps instead of images, where the downstream models shall to be trained for adapting the compressed features.
Additionally, various researches~\cite{choi2018high,huang2021visual,choi2020task,cai2021novel,zhang2021just} have elevated traditional codecs by introducing an independent task-specific feature encoding stream~\cite{veselov2021hybrid,chao2015keypoint}, or utilizing hand-crafted structure maps~\cite{akbari2019dsslic,hu2020towards,duan2022jpd}.
Furthermore, some works~\cite{bai2022towards,yang2020discernible,duan2020video,choi2022scalable11,ge2024task}directly incorporate the downstream task loss.
Recently, several works~\cite{dubois2021lossy,feng2022image} have adopted self-supervised loss as the data-driven semantic source.
But, they still require task-specific training, limiting their practicality.
%Most above methods require to be trained or fine-tuned with task-specific annotations.
% which are infeasible due to the laborious video annotation procedure and the difficulties of fine-tuning video networks.

\textbf{Unsupervised Video Semantic Compression (UVSC).}
Recently, the concept of UVSC has emerged for catering to real-world scenarios, where compressed videos from a unified codec are directly used for various downstream tasks~\cite{lin2019tsm,tian2019video,tian2020self,bertasius2021space,wang2021tdn,tian2022ean,li2023uniformer,li2022mvitv2,wang2022internvideo,zhang2021bytetrack,yan2021dehib,yi2021attention,che2021adversarial,duan2022develop,cao2023observation,cheng2022xmem,tian2018ban,chen2024gaia,chen2024cross,wang2023look,yan2021dehib,dhbe_asiaccs_2023} without any task-specific training.
%Previous video compression approaches, including traditional~\cite{bross2021overview}, learnable~\cite{li2023neural}, ROI-based~\cite{cai2021novel}, and structure-enhanced~\cite{duan2022jpd} methods, all fall under the broad UVSC concept, since their decoded videos can be directly fed into downstream modules for performing AI tasks.
%But, they are sub-optimal, due to not particularly preserving the semantic information or only relying on hand-crafted strategies (\textit{e.g.}, ROI, segmentation map).
Recent works~\cite{tian2024coding,tian2023non,tian2024smc++} propose to optimize a semantics-preserving objective such as contrastive learning~\cite{he2020momentum} or MAE~\cite{he2022masked}\cite{tong2022videomae} loss, as well as another photo-realism regularization item such as GAN~\cite{goodfellow2020generative}, to compress the video into a semantic-full yet high-quality video at very low bitrate.
% introduces a baseline UVSC framework with two loss terms: MAE loss for learning semantics and another one for photo-realism. 
Nonetheless, the semantics obtained by a single self-supervised learning loss is still not sufficiently rich.
In this work, we leverage the pre-trained VFMs to absorb much richer semantics.

\textbf{Visual Foundation Models.}
%ConvNets~\cite{lecun1989backpropagation}, such as VGGNet~\cite{simonyan2014very}, ResNet~\cite{he2016deep}, InternImage~\cite{wang2023internimage}, and SlowFast~\cite{feichtenhofer2019slowfast}, which are pre-trained on large-scale image/video classification datasets~\cite{deng2009imagenet,carreira2017quo}, have long served as the standard visual foundation models (VFMs), forming the backbone for various downstream tasks.
Recently, numerous works have combined the powerful modeling capabilities of Transformers~\cite{vaswani2017attention,dosovitskiy2020image,liu2021swin,zhaovideoprism} with self-supervised learning methods, resulting in a series of powerful VFMs for both images~\cite{xie2021self,he2022masked,oquab2023dinov2} and videos~\cite{tong2022videomae,wang2023videomae,wang2022internvideo}.
For the first time, we explore how to harness current VFMs for the unsupervised video semantic compression problem.

\textbf{Inter-Frame Entropy Models.}
Traditional~\cite{sullivan2012overview,bross2021overview} and state-of-the-art learnable codecs~\cite{hu2022coarse,li2023neural} typically use motion to model the inter-frame redundancy.
However, explicit pixel (or region) movement-based motions are often suitable for
modeling the temporal dynamics of low-level features, instead of more abstract high-level semantic features.
While a few methods~\cite{liu2020conditional,mentzer2022vct,yang2022advancing} directly predict entropy model parameters or features for current frame, their models are static.
In contrast, our trajectory-based entropy model is content-adaptive and dynamic.

\vspace{-4mm}
\section{Approach}
\vspace{-2mm}
\label{sec:method}

\begin{figure*}[!t]
	%	\vspace{-3mm}
	\centering
	\includegraphics[width=12.5cm]{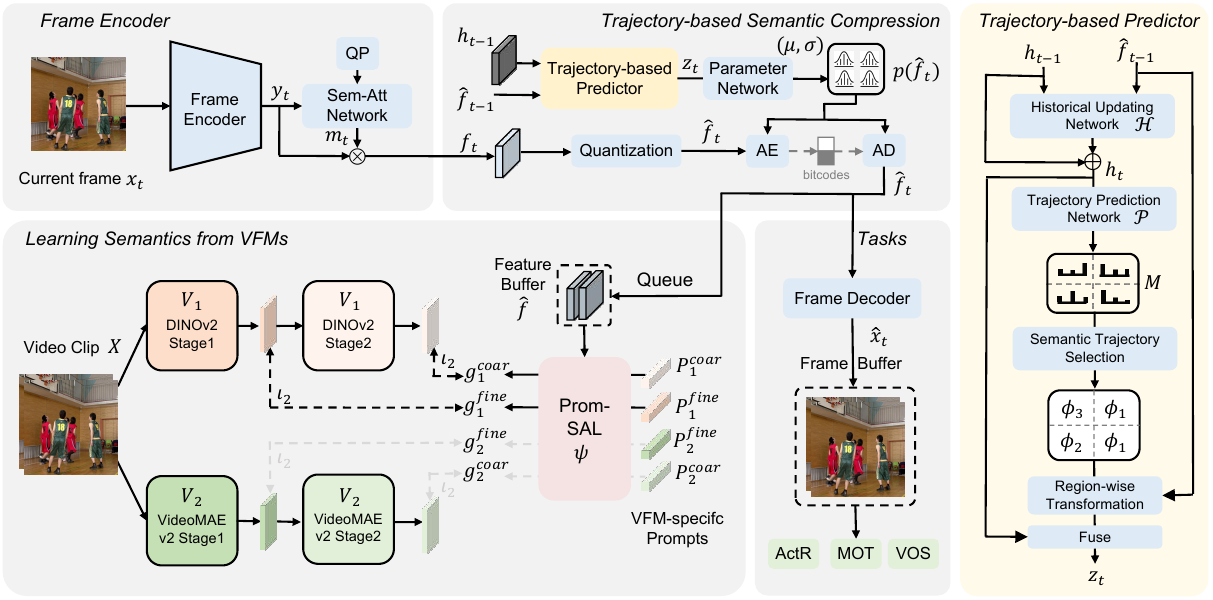}
	\vspace{-4mm}
	\caption {
		Overview of Free-VSC framework, which learns to absorb rich semantics from multiple VFMs into the compression procedure.
		A prompt-based semantic alignment layer (Prom-SAL) is introduced to flexibly align the compressed video feature $\hat{f}$ to the semantic space of VFMs.
		We also propose a trajectory-based entropy model for efficiently compressing the inter-frame semantic redundancy.
		We illustrate two VFMs ($V_1$ and $V_2$) and three semantic trajectories for simplicity, although more VFMs and trajectories can be applied in our approach.
	}
	\vspace{-4mm}
	\label{fig:framework}
\end{figure*}

In this paper, we propose a new unsupervised video semantic coding framework termed Free-VSC (Figure~\ref{fig:framework}),
which absorbs rich semantic knowledge from multiple VFMs.
The main learning objective is to minimize the discrepancy between the compressed and original videos in a VFMs-aligned semantic space.
Besides, a semantic trajectory-based entropy model is proposed for more efficient inter-frame semantic compression.

%\vspace{-2mm}
%\subsection{Framework Overview}
%\vspace{-2mm}
\textbf{Encoding.} Let an input video $X = \{x_1, x_2, ..., x_t, ..., x_T\}$, where $T$ denotes the video temporal length,
each frame $x_t$ is first encoded and quantized as feature $\hat{f}_t$ by a frame encoder on the sender side.
Then, based on the historical feature information $\{\hat{f}_{1}, ..., \hat{f}_{t-1}\}$, the possible semantic trajectory of each region is predicted as a distribution of non-linear transformations.
The past frame feature $\hat{f}_{t-1}$ is transformed along the trajectories, serving as the context feature for calculating the entropy model $p(\hat{f}_t)$.
Finally, an arithmetic encoder (AE) uses $p(\hat{f}_t)$ to encode $\hat{f}_t$ as bitcodes.

\textbf{Decoding.} On the recipient side, all past quantized features $\{\hat{f}_{1}, ..., \hat{f}_{t-1}\}$ have been received and are accessible, so that the entropy model $p(\hat{f}_t)$ can be precisely reconstructed without transporting any side information.
The bitcodes can be decoded back into $\hat{f}_t$ with $p(\hat{f}_t)$ by an arithmetic decoder (AD),
which are further restored as a video frame $\hat{x}_t$ by a frame decoder network.
The decoded frames are fed into various video analysis modules,
such as video action recognition (ActR), multiple object tracking (MOT) and video object segmentation (VOS) models.

\textbf{Learning.} During the optimization of the compression system, we randomly select one VFM $V_n$ from the candidate VFM set as the semantic guidance in each optimization step.
Compared to simultaneously training with multiple VFMs, randomly activating a single VFM
% avoids balancing semantic features from different VFMs, as well as 
enables scaling VFM number more easily, due to the substantially saved GPU memory.
Upon receiving the compressed video semantic feature $\hat{f} = \{\hat{f}_{1}, ..., \hat{f}_{T}\}$, we employ a prompt-based semantic alignment layer (Prom-SAL) and the VFM $V_n$-specific prompt $P_n$ to align it within the semantic space of $V_n$, thereby generating the VFM-aligned compressed semantic feature $g_n$.
Then, a simple $\ell2$ loss is utilized to minimize the discrepancy between $g_n$ and the original video feature extracted by $V_n$.
The learning procedure is conducted in a single stage, \textit{i.e.}, both the encoder, decoder and Prom-SAL are simultaneously optimized in an end-to-end fashion.

The details of each component are elaborated as follows.

\vspace{-2mm}
\subsection{Frame Encoder and Decoder Networks}
\vspace{-2mm}
We employ a frame encoder network to transform the frame $x_t \in \mathbb{R}^{3 \cdot H \cdot W}$ into deep features $y_t \in \mathbb{R}^{256 \cdot \frac{H}{32} \cdot \frac{W}{32}}$.
To adapt to various compression quality parameters (QPs), we further append a QP-conditioned semantic attention network (Sem-Att Net). This network generates a spatial-channel attention mask $m_t$ for masking the information according to different QP values.
The final semantic feature is obtained by multiplying the original feature with the attention mask, denoted as $f_t = y_t \odot m_t \in \mathbb{R}^{256 \cdot \frac{H}{32} \cdot \frac{W}{32}}$, where $\odot$ represents the element-wise multiplication operation. The resulting feature $f_t$ is quantized as $\hat{f}_t$ for transmission.
We mention that using QP as the conditional variable to adjust feature bitrate is a common practice in variable-rate compression~\cite{yang2020variable}.

The encoder frame network consists of five stages. The initial stage includes a convolution layer with stride size two, followed by a residual block. Each of the subsequent four stages comprises a convolution layer with stride size two, succeeded by two inverted residual blocks~\cite{sandler2018mobilenetv2} and two residual blocks. The Sem-Att Net is simply composed of three convolution layers with kernel size three, followed by a sigmoid function.

%\subsection{}
The frame decoder comprises five stages, progressively upsampling low-resolution features $\hat{f}_t$ to full-resolution frames. The first stage includes a single upsampling layer implemented with transpose convolution.
The second/third stages include eight/four residual blocks followed by a upsampling layer, while the last two stages include only one residual block for efficient computation.
%The /third/fourth/fifth stages include six/two/one/one/one 
%The second stage consists of six residual blocks with a hidden channel number $c$, followed by a upsampling layer. The third stage follows a similar pattern with six residual blocks, but with a hidden channel number \(\frac{c}{2}\).
%The fourth stage includes two residual blocks with a hidden channel number \(\frac{c}{4}\) followed by a upsampling layer, while the fifth stage consists of one residual block with a hidden channel number \(\frac{c}{8}\) followed by a upsampling layer.
%The majority of residual blocks and parameters are allocated to second/third stages and low-resolution feature spaces, enabling efficient calculation.
%specifically \(\frac{1}{16}\) or \(\frac{1}{8}\) of the original resolution.

%We adopt a fully convolutional network by stacking eight residual layers followed by pixel-shuffle operations~\cite{shi2016real} to reconstruct the frame $\hat{x}_t$ from the semantic feature $\hat{f}_t$.

\vspace{-2mm}
\subsection{Trajectory-based Semantic Compression}
\vspace{-2mm}
As shown in the right side of Figure~\ref{fig:framework}, we first recurrently aggregate the past information as the hidden state $h_t$, which is of the same size as the frame feature $f_t$.
Then, we predict the context feature $z_t$ of the current frame feature $\hat{f}_t$ via dynamically predicting the region-wise semantic trajectories.
% in the latent space.
%use a trajectory prediction network $\mathcal{P}$ to predict the local trajectories for each location.
% the region-wise distribution map of  semantic trajectories $M_t \in \mathbb{R}^{N \cdot h \cdot w}$, where $N$ is the number of possible local trajectories.
%The local trajectory is obtained by querying a trajectory pool with $M_t$. 
%Finally, the historical information is transformed by the local trajectories, producing the context feature $z_t$ of current frame.
Finally, conditioned on $z_t$, a contextual entropy model is employed to compress $\hat{f}_t$.
The detailed procedure is expanded as follows.
%Finally, a parameter network is employ to predict the parameters of probability model for coding current frame semantics.

\textbf{Historical State Aggregation.}
Formally, given the previously aggregated state $h_{t-1}$ and the previous frame feature $\hat{f}_{t-1}$, we employ a historical updating network $\mathcal{H}$ to combine them, resulting in a new state $h_{t} = h_{t-1} + \mathcal{H} (h_{t-1},\hat{f}_{t-1} )$. Subsequently, $h_{t}$ is recurrently used as the historical information in the coding loop of the next frame.
The network $\mathcal{H}$ comprises three convolutions with kernel size three and LeakyReLUs~\cite{xu2015empirical}.

\textbf{Trajectory-based Context Prediction.}
Rather than directly predicting context from $h_{t}$ using a static transformation, which can struggle with diverse contents across different frames or videos, we propose to (1) first dynamically predict the possible semantic trajectory of each region, and (2) then synthesize the context by transforming the previous feature with trajectories in a region-wise manner.
We select the dynamic trajectories from a few possible candidates instead of directly predicting their parameters for easier optimization, following previous work~\cite{chen2020dynamic}.
The possible trajectories are defined as a pool of non-linear transformations $\{\phi_1, \phi_2, ..., \phi_S\}$, where $S$ denotes the total number.

% of available transformations.
%We define semantic trajectories in discrete space, following the discrete nature of image semantics~\cite{tan2021diverse}, and reducing the solution space.
%The detailed procedure are given as followings.

We employ a trajectory prediction network to predict the region-wise trajectory probability map $M_t$.
For each spatial position $o = \{o_x,o_y\}$, the transformation with the largest probability
% \scalebox{0.8}{$\mathcal{T} = \underset{s}{\mathrm{argmax}} M_t[s,o]$} 
is selected as the semantic trajectory $\phi_s$.
% where\scalebox{0.8}{ $
%	s =  \underset{s}{\mathrm{arg~max}}~ M_t[s,o]
%	$}. 
$\phi_s$ is used to transform the previous feature $\hat{f}_{t-1}[o]$ to predict the future feature $f'_t[o] = \phi_s (\hat{f}_{t-1}[o])$.
% of which each position $o$ feature $f'_t(o) = \phi_{} $
%To enable the end-to-end training, we calculate the above `select than transform' procedure as `sum over multiplication'.
% \textit{i.e.},
%$
%%\begin{equation}
%%\end{equation}
%f'_t [o] = \sum_{s=1}^{S} \phi_s ( \hat{f}_{t-1}[o] ) \odot  M_t[s,o] ,
%%	\end{equation}
%$
%where
%$\odot$ represents the channel-wise broadcasting multiplication operation.
Finally, we utilize a $\operatorname{Fuse}$ operation to enhance the predicted future feature $f'_t$ with the long-term information, producing the current frame coding context, \textit{i.e.}, $z_t = \operatorname{Fuse}(f'_t \oplus h_t)$, where $\oplus$ denotes the channel-wise concatenation operation.

As for network details, the trajectory prediction network consists of three convolution layers with kernel size three, followed by a $softmax$ operation.
$\operatorname{Fuse}$ comprises three residual blocks, each of which includes one convolution with kernel size three.
%All convolution layers are followed by LeakyReLU non-linearity~\cite{xu2015empirical}.

\textbf{Context-based Compression.}
Given the context feature $z_t$, we employ three stacked convolutions with kernel size three, followed by a three-layer multiple-layer-perceptron (MLP), to predict two sets of parameters, namely mean $\mu$ and variance $\sigma$, which are used to construct the Gaussian distribution model $p(\hat{f}_t) \sim \mathcal{N}(\mu,\sigma)$.
$\hat{f}_t$ is compressed by an arithmetic encoder with $p(\hat{f}_t)$~\cite{balle2018variational}.
The sizes of $\mu$ and $\sigma$ are the same as that of $\hat{f}_t$.
% uses this model to encode .

\vspace{-2mm}
\subsection{Learning Rich Semantics from VFMs}
\vspace{-2mm}
Given the compressed video semantic feature $\hat{f} = \{\hat{f}_1, ..., \hat{f}_T\}$, we propose to align it with pre-trained VFMs, within the shared semantic space of VFMs.
% that are denoted as $\{V_1, V_2, ..., V_N \}$.
% to measure semantic distortion. 
%Here, $N$ represents the number of the adopted VFMs. 
We compute the feature distance in this aligned space as the semantic distortion term. In each iteration, a VFM is randomly selected as the semantic source, avoiding balancing different VFMs. Additionally, this approach involves only one VFM in each forward-backward procedure, ensuring memory efficiency.
In the following, we provide a detailed description of the semantic learning procedure.
%in one training step.

Given the selected VFM $V_n$, we input the original video $X$ into it, resulting in two groups of feature maps representing fine-level semantics (in shallow layer) and coarse-level semantics (in deep layer), denoted as, $V_n(X)^{fine}$ and $V_n(X)^{coar}$.
The adoption of multi-level features is for catering to the downstream tasks of various granularities, \textit{i.e.}, global tasks such as recognition rely mostly on coarse-level features, while local tasks such as tracking/segmentation require both coarse- and fine-level features.
%We refer to the features from the shallow layer as "fine semantics" because they typically contain more detailed information about objects.
These two-level features are separately normalized by dividing the moving average of their $\ell2$ norm. This normalization step eliminates the magnitude differences among features obtained from different VFMs.
For simplifying the notations, we denote the normalized features also as $V_n(X)^{fine}$ and $V_n(X)^{coar}$. 

Then, we introduce the prompt-based semantic alignment layer (Prom-SAL) to align our compressed semantics to the semantic features from different VFMs.
% within the shared semantic space of VFMs. 
Prom-SAL mainly includes a spatial-temporal Transformer $\psi$.
Motivated by the conclusion in VPT~\cite{jia2022visual}, \textit{i.e.}, input prompts can effectively tune Transformer's behavior towards a specific learning target, we also equip $\psi$ with a set of the VFM-specific prompt tokens.
We denote the prompts for $V_n$ as $\{P_n^{fine} \in \mathbb{R}^{1024}, P_n^{coar} \in \mathbb{R}^{1024}\}$, which are learnable and initialized with values drawn from a Gaussian distribution.
%  to transform the received semantic features $\hat{f}$ into the semantic space of this VFM,
Given the inputs above, Prom-SAL produces the VFM $V_n$-aligned features $g_n^{fine} = \psi(P_n^{fine},\hat{f})$ and $g_n^{coar} = \psi(P_n^{coar},\hat{f})$.
%The alignment function $\psi$ is also shared when aligning coarse or fine level features, for reducing the parameter number.
Then, the semantic distortion can be formulated as:
% loss in the VFMs-shared semantic space is given by:
\vspace{-1mm}
\begin{equation}
%	\scalebox{0.9}{$
		\begin{aligned}
			\mathcal{D}_{sem} =  \ell_2(g_n^{fine}, V_n(X)^{fine})+\ell_2(g_n^{coar}, V_n(X)^{coar}),
		\end{aligned}
%	$
%	}
	\vspace{-1mm}
\end{equation}
where $\ell_2$ denotes the mean square error (MSE) loss.

%\subsection{Optimization of Framework}\label{method_learning}
Finally, the trade-off objective between the bitrate and the distortion of the semantic feature can be formulated as,
\vspace{-1mm}
\begin{equation}
	\scalebox{1}{$
		\mathcal{L}_{RDsem} =  QP \cdot \frac{1}{T}\sum_{t=1}^{T} R(\hat{f}_t) + \mathcal{D}_{sem},
		$
	}
	\vspace{-1mm}
\end{equation}
where the compression quality parameter $QP$ is randomly sampled from $[0.5,4]$, aiming to train a unified compression model for variable bitrates.
$T$ denotes the length of the training clip.
$R(\hat{f}_t)$ is the entropy~\cite{balle2018variational} of $\hat{f}_t$ with the distribution $p(\hat{f}_t)$.
Following previous works~\cite{tian2023non}\cite{esser2021taming}, we also enforce the image perceptual loss~\cite{zhang2018unreasonable} $\mathcal{L}_{percep}$ and Generative Adversarial Network (GAN)~\cite{goodfellow2020generative} loss $\mathcal{L}_{GAN}$ on the decoded frame $\hat{x}_t$, to ensure the photo-realism of the decoded frames. Finally, the whole loss function of the system is given by $\mathcal{L} =   \mathcal{L}_{RDsem}+  \mathcal{L}_{percep}  +  \mathcal{L}_{GAN}$.
%\begin{equation}
%\mathcal{L} =   \mathcal{L}_{RDsem}+  \mathcal{L}_{percep}  +  \mathcal{L}_{GAN}.
%\end{equation}
The architecture and loss function of GAN follow that of PatchGAN~\cite{isola2017image}.

\vspace{-4mm}
\section{Experiments}
\vspace{-2mm}
\subsection{Experimental Setup}
%\vspace{-2mm}
\textbf{Training Dataset.} 
Following the previous works \cite{wu2018video,tian2023non}, we use 60K videos from the Kinetics400 dataset~\cite{carreira2017quo} during the training stage.
For data augmentation, the video sequences are randomly flipped and cropped
into $256 \times 256$ patches.

\textbf{Evaluation Datasets.}
Our evaluation protocol rigorously follows \cite{tian2023non}.
For {action recognition} task, we evaluate it on {UCF101}~\cite{soomro2012ucf101}, {HMDB51}~\cite{kuehne2011hmdb}, {Kinetics}~\cite{carreira2017quo}, and {Diving48}~\cite{li2018resound}.
%We pre-downsample the shortest side of them to 256 pixels and crop the video to the size 224$\times$224 before the coding procedure.
For {multiple object tracking (MOT)} task, we evaluate it on {MOT17}~\cite{milan2016mot16}.
%We adopt the original MOT17 dataset of resolution 1920$\times$1080 because many tracking methods require high-resolution inputs.
For {video object segmentation (VOS)} task, we evaluate it on {DAVIS2017}~\cite{pont20172017}.

\textbf{VFM Details.}
We have adopted three representative self-supervised VFMs:
-VideoMAEv2~\cite{wang2023videomae}, utilizing ViT-B network~\cite{dosovitskiy2020image} and trained with MAE loss~\cite{he2022masked}.
-DINOv2~\cite{oquab2023dinov2}, based on ViT-B network and trained using a combined objective, incorporating DINO~\cite{caron2021emerging}, iBOT~\cite{zhou2021ibot}, and SwAV~\cite{caron2020unsupervised}.
-SSL-Swin~\cite{xie2021self}, leveraging the Swin-B network~\cite{liu2021swin} and trained with a combined objective of MoCo v2~\cite{chen2020improved} and BYOL~\cite{grill2020bootstrap}.
These models cover mainstream self-supervised learning paradigms, such as masked image/video modeling~\cite{he2022masked} and contrastive learning~\cite{he2020momentum}. Additionally, they incorporate two popular backbone architectures: non-hierarchical (ViT) and hierarchical (Swin).
The `fine' and `coarse' layers in all VFMs correspond to the `4th' and `8th' attention blocks.
%roughly the one-third and two-thirds depth locations. 
We avoid using very deep layers, as they are biased towards the used self-supervised learning objectives, potentially limiting generalizability.

\textbf{Implementation Detail.}
The I frame is encoded with the channel-wise auto-regressive entropy model~\cite{minnen2020channel}, where the channel group number is set to eight.
For the first P frame, the historical state $h_{t-1}$ of the entropy model is initialized as the I frame feature.
The Transformer in Prom-SAL is instantiated as six attention blocks of ViT, where input channel number, MLP hidden channel number and attention header number are set to 1024, 2048 and 32, respectively. $sin/cos$ positional embedding is adopted.
%For each VFM, $\psi$ is also appended with a VFM-specific linear layer to align the tensor dimensions between the features from $\psi$ and VFM.
The number of possible trajectories $N$ is empirically set to 5.
The quality parameter (QP) set is $\{0.5,2,3,4\}$.
The video clip length $T$, is fixed at eight during training.
%Our framework is optimized in a single stage and more concise than recent learnable methods~\cite{hu2022coarse,li2023neural}, which typically involve more than two stages.
The initial learning rate is set to 1e-4 and is reduced by a factor of ten at 500,000 steps. The total number of training steps is 600,000.
The mini-batch size is 16.
We utilize the Adam optimizer~\cite{kingma2014adam} implemented in PyTorch~\cite{paszke2019pytorch} with CUDA support. The values of $\beta1$ and $\beta2$ are set to 0.9 and 0.999, respectively.
The entire training process takes approximately five days on a machine equipped with four NVIDIA GeForce RTX 4090 GPUs.

%\subsection{Experimental Setting}

\textbf{Baseline Methods.}
%Beyond the methods in previous benchmark~\cite{tian2023non}, \textit{i.e.},  {FVC}~\cite{hu2021fvc}, {PLVC}~\cite{yang2021perceptual} and , we incorporate more representative methods into the benchmark.
We compare our method with traditional codecs, namely {HEVC}~\cite{tomar2006converting} and {VVC}~\cite{bross2021overview,VVenC}, learnable codecs, namely {FVC}~\cite{hu2021fvc}, {PLVC}~\cite{yang2021perceptual} and {DCVC-DC}~\cite{li2023neural}, as well as recent semantics-oriented coding methods, namely {ROI}~\cite{cai2021novel}, {JPD-SE}~\cite{duan2022jpd} and {SMC}~\cite{tian2023non}.
All approaches are evaluated on low delay P mode (LDP) with group-of-picture(GOP) size 10 for a fair comparison.

\textbf{Task Models.}
%Our downstream models follow the previous benchmark~\cite{tian2023non} for fair comparisons.
For the {action recognition} task, we evaluate on {TSM}~\cite{lin2019tsm}, {SlowFast}~\cite{feichtenhofer2019slowfast}, and {TimeSformer}~\cite{bertasius2021space} networks.
The model weights are provided by MMAction2~\cite{2020mmaction2}.
For the {MOT} task, we adopt {ByteTrack}~\cite{zhang2021bytetrack}. The model weight is provided by MMTracking~\cite{mmtrack2020}.
For the {VOS} task, we adopt {XMem}~\cite{cheng2022xmem} and the official model.
%, of which the model weights is released by authors.
%We directly feed the decoded videos by our framework into these official models, without any model fine-tuning.

\textbf{Evaluation Metrics.}
We use bpp (bit per pixel) to measure the average number of bits used for one pixel in each frame.
For the {action recognition} task, we adopt the Top1 accuracy metric.
For the {MOT} task, we adopt MOTA (multiple object tracking accuracy)~\cite{kasturi2008framework}, MOTP (multiple object tracking precision), FN (false negative detection number) and IDF1 metrics. 
For the {VOS} task, Jaccard index $\mathcal{J}$, contour accuracy $\mathcal{F}$, the average of $\mathcal{J}$ and $\mathcal{F}$ ($\mathcal{J\&F}$), and contour recall $\mathcal{F}$-$Recal$ are adopted.
%We also report the .

\vspace{-4mm}
\subsection{Experimental Results}
\vspace{-2mm}

\textbf{Action Recognition.}
As shown in Table~\ref{tab:benchmark_ar}, when using TSM as the action recognition network, our method demonstrates superior performance to all other methods, \textit{i.e.}, the recent learnable method DCVC-DC~\cite{tan2021diverse} and semantic codec SMC~\cite{tian2023non}.
On lower bitrate settings, such as HMDB51@0.02bpp and Diving48@0.05bpp, we observe a 27.02\% and 19.30\% performance gain over DCVC-DC, along with 10.12\% and 12.01\% accuracy enhancements over SMC.
In higher bitrate settings, our method also maintains superiority, \textit{e.g.}, outperforming SMC by 2.70\%, 2.57\%, and 7.67\% on UCF101@0.04bpp, Kinetics400@0.06bpp, and Diving48@0.07bpp settings, respectively.
Moreover, when compared to other hand-crafted semantic representation-based methods ROI~\cite{cai2021novel} and JPD-SE~\cite{duan2022jpd}, our approach achieves more than 10\% performance gain on the large-scale Kinetics400@0.06bpp setting.
\begin{figure*}[!t] 
%	\vspace{-2mm}
	\centering
	\renewcommand\arraystretch{0.0}
	\newcommand{\widthscalefive}{0.24}
	\tabcolsep = 0.6mm
	\scalebox{1}{
		\begin{tabular}{cccc}
			\includegraphics[width=\widthscalefive \textwidth]{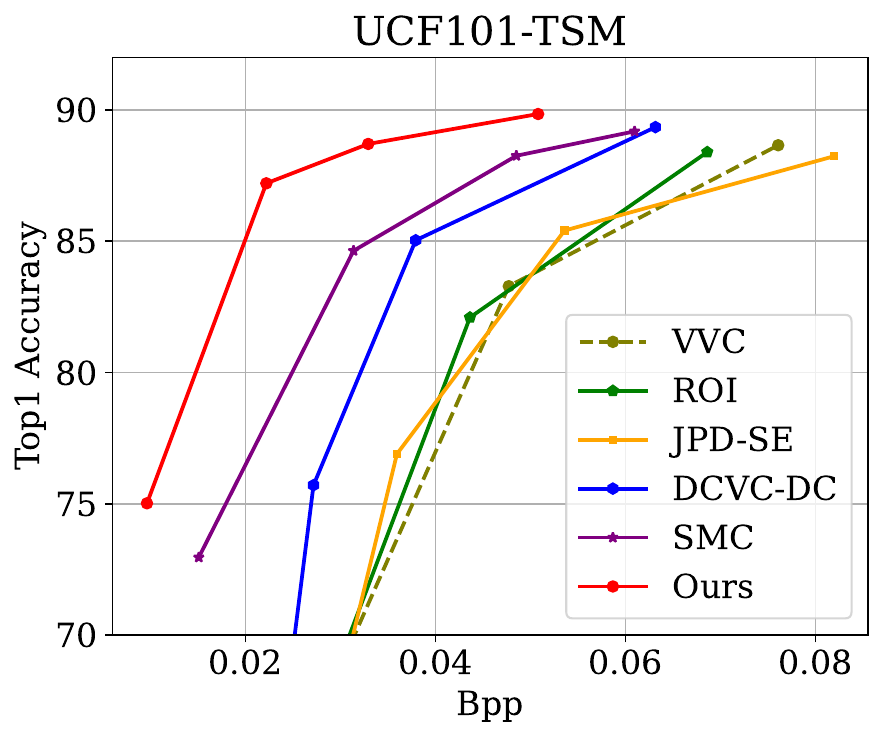} &
			\includegraphics[width=\widthscalefive \textwidth]{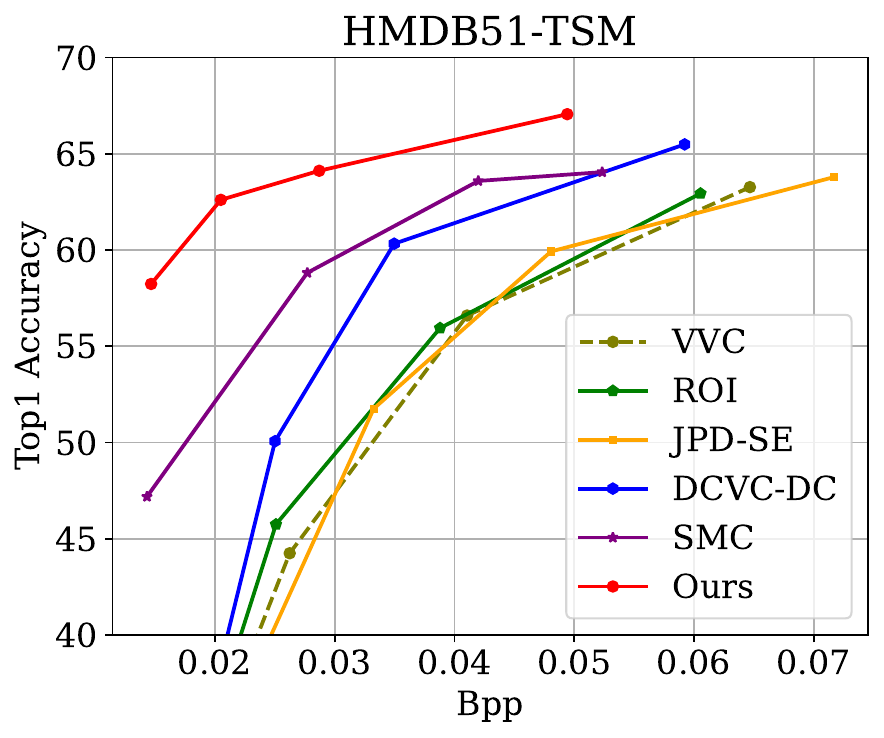} &
			\includegraphics[width=\widthscalefive \textwidth]{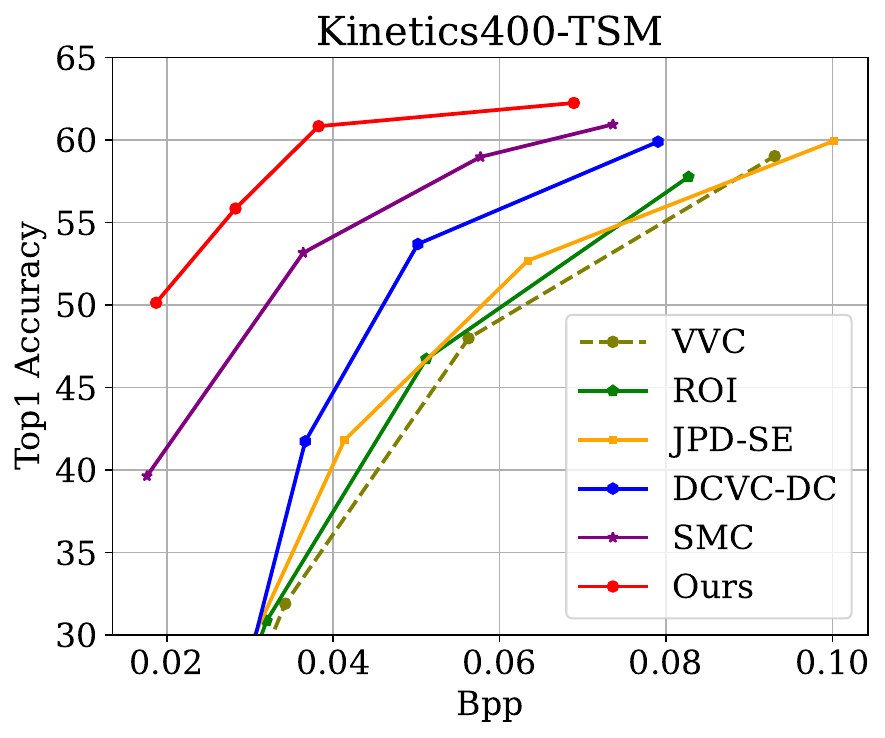} &
			\includegraphics[width=\widthscalefive \textwidth]{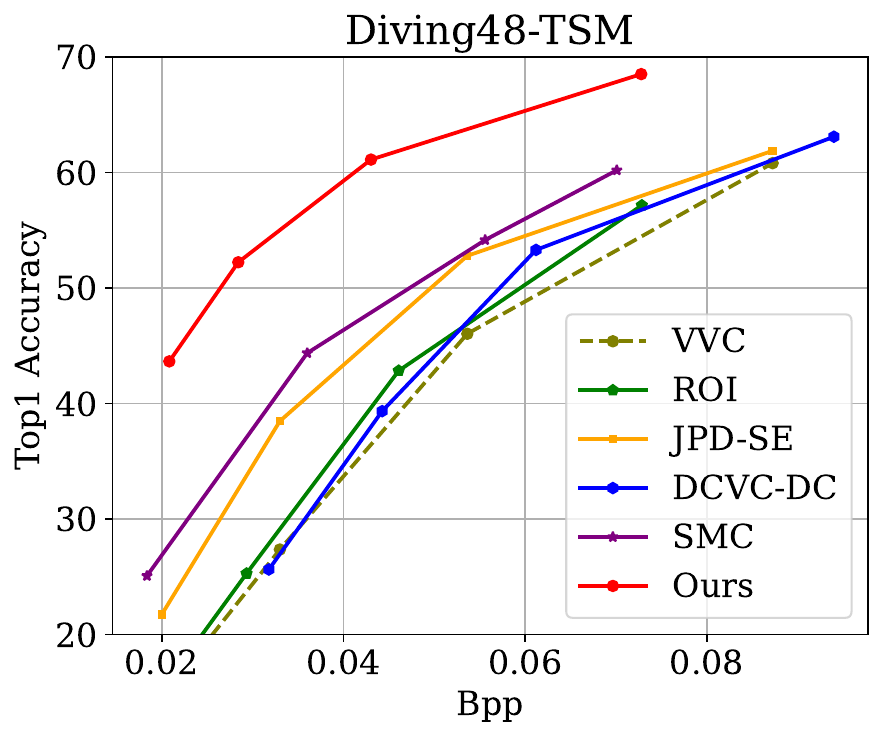} \\
			\includegraphics[width=\widthscalefive \textwidth]{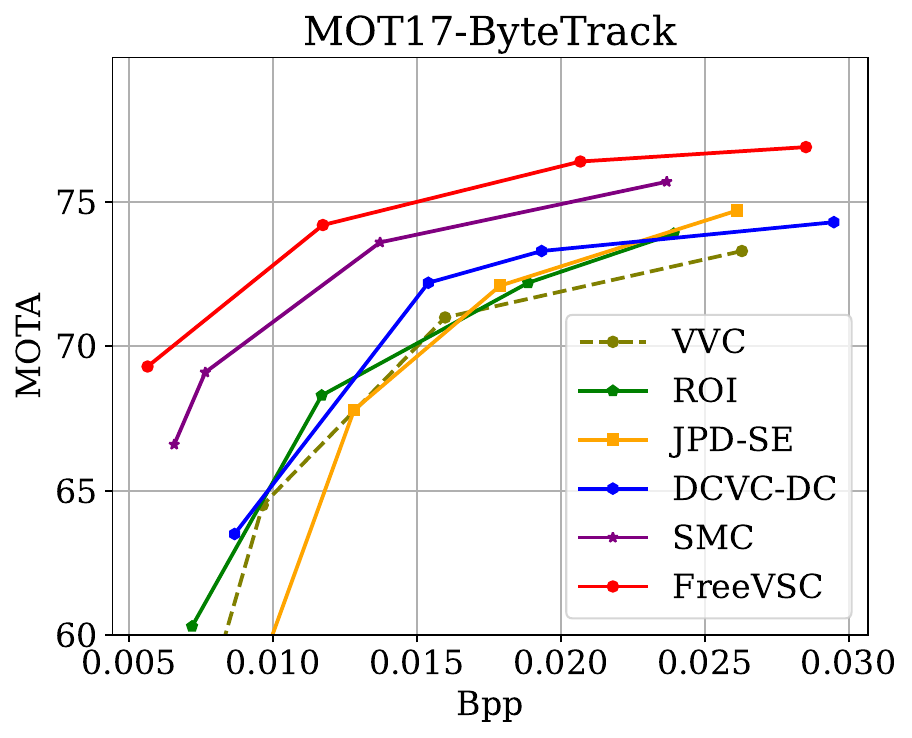} &
			\includegraphics[width=\widthscalefive \textwidth]{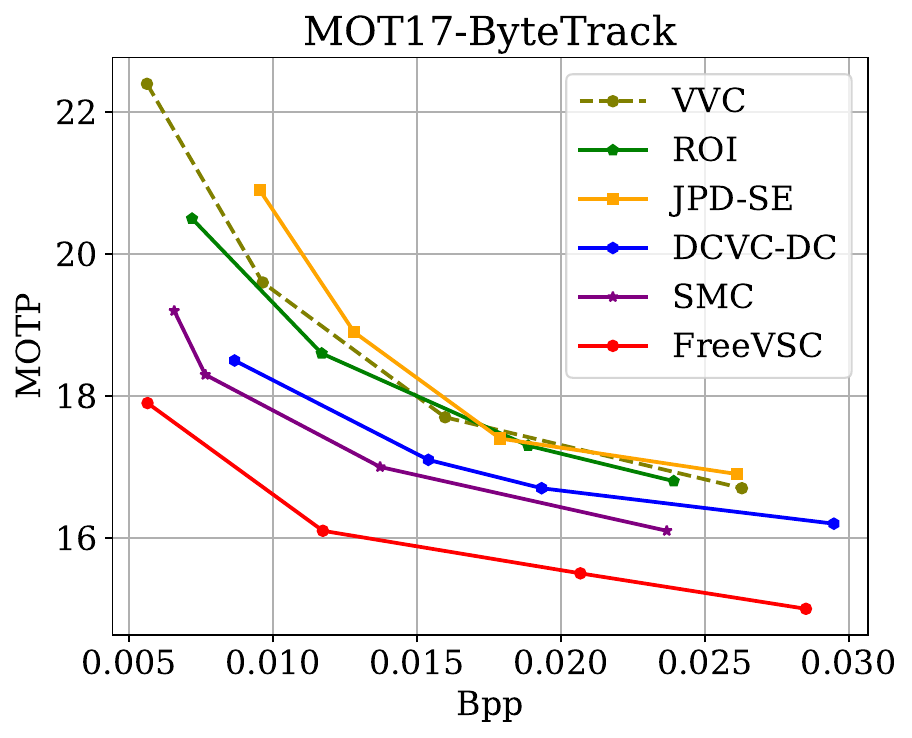} &
			\includegraphics[width=\widthscalefive \textwidth]{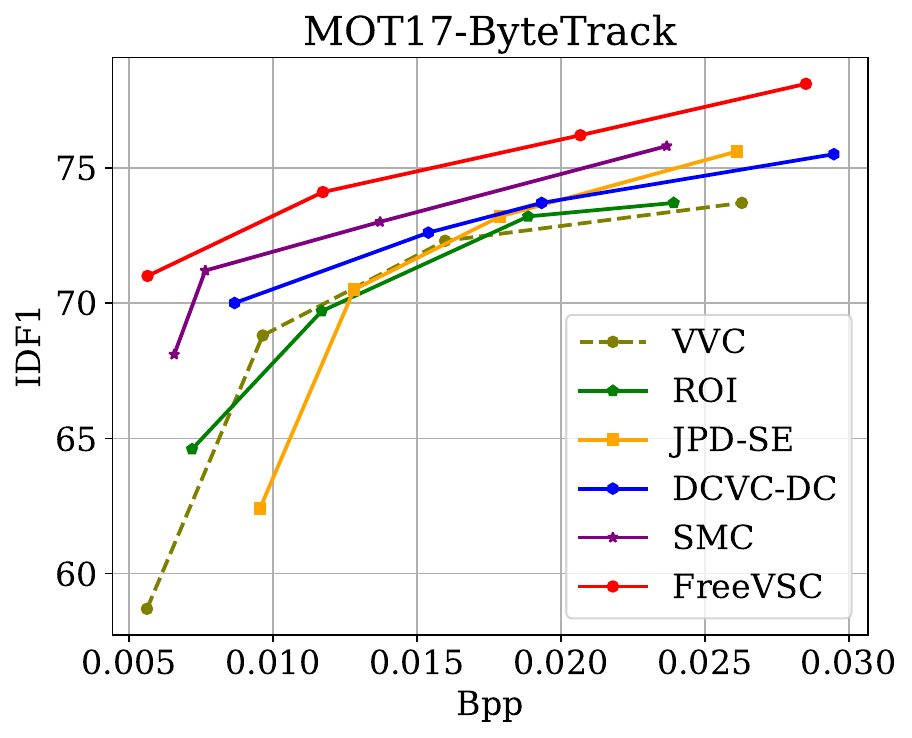} &
			\includegraphics[width=\widthscalefive \textwidth]{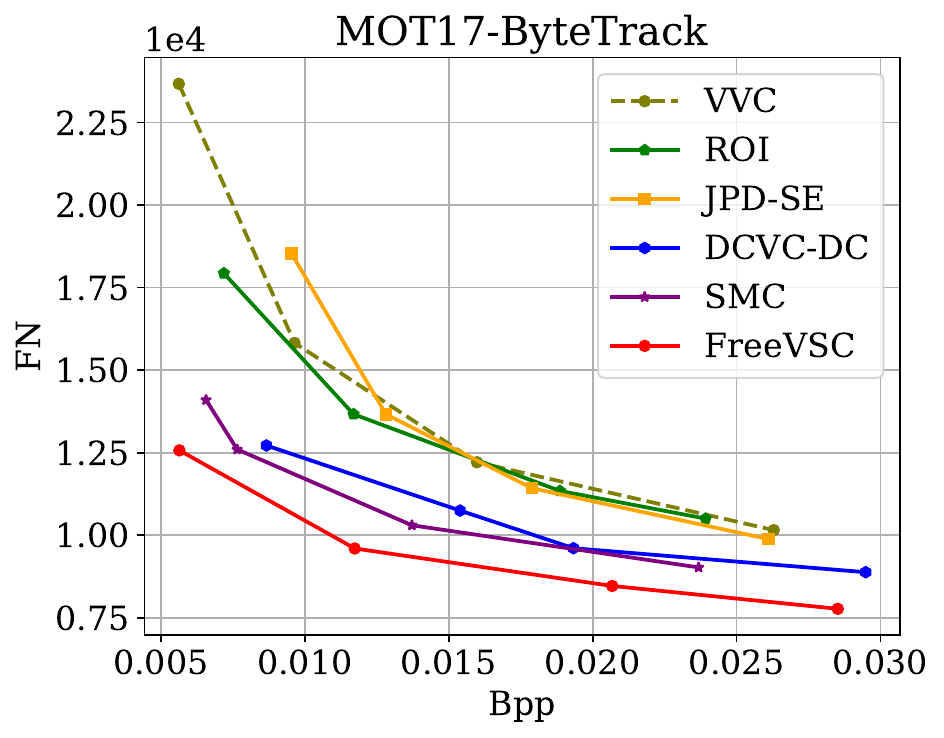} \\
			\includegraphics[width=\widthscalefive \textwidth]{\detokenize{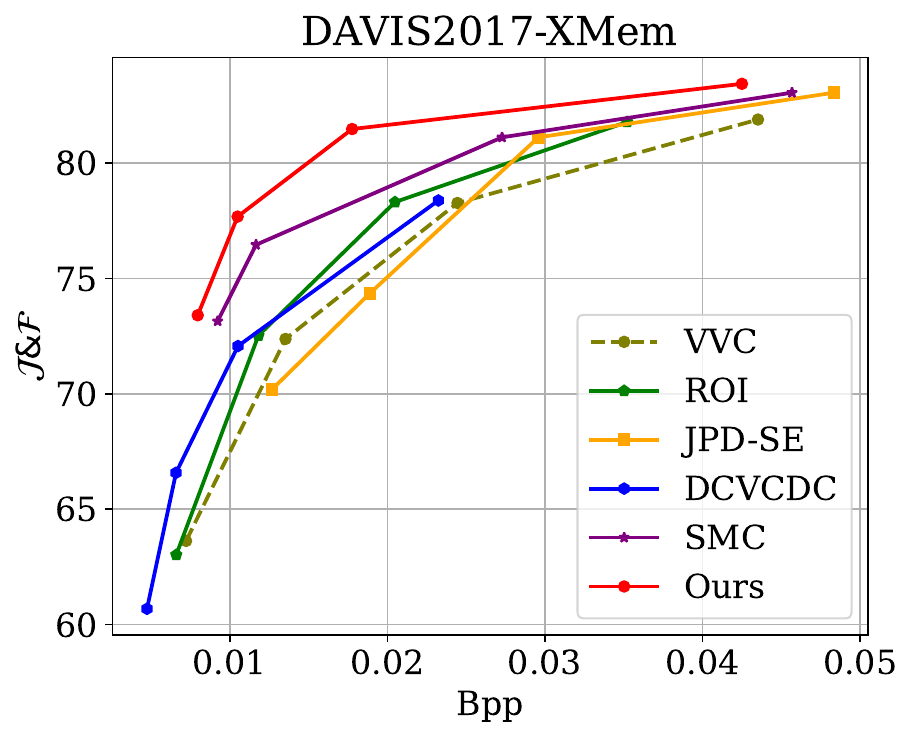}} &
			\includegraphics[width=\widthscalefive \textwidth]{\detokenize{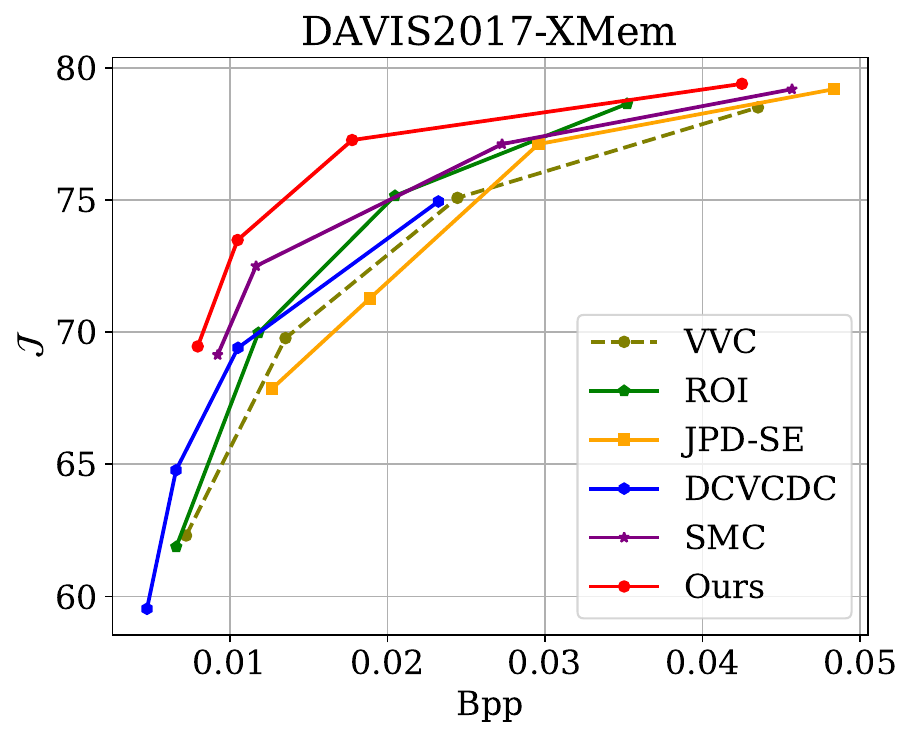}} &
			\includegraphics[width=\widthscalefive \textwidth]			   {\detokenize{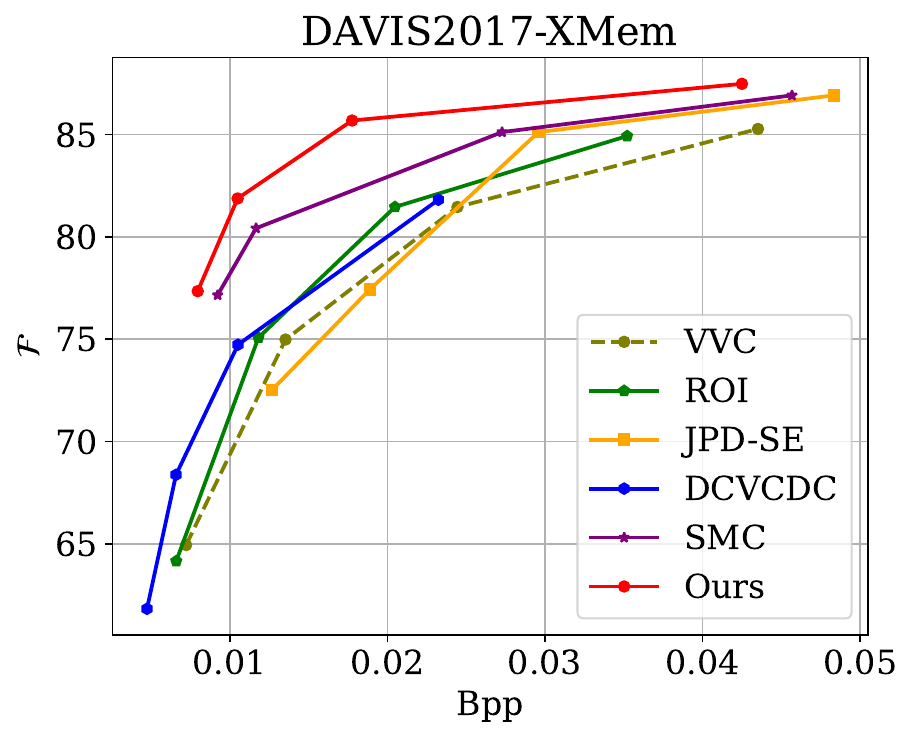}} &
			\includegraphics[width=\widthscalefive \textwidth]{\detokenize{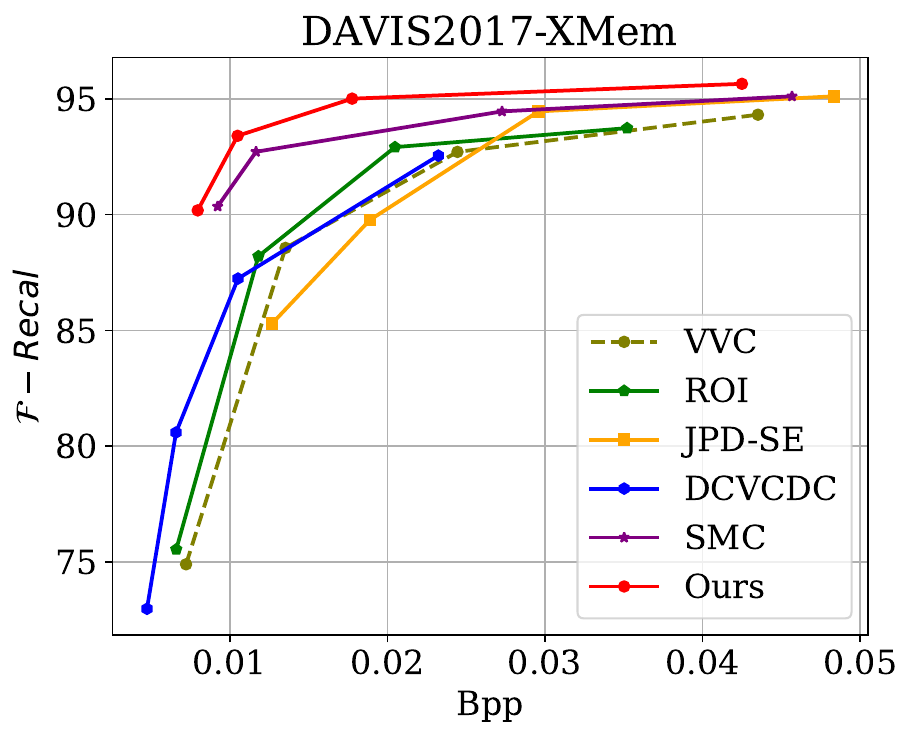}} 
			%				 \\
		\end{tabular}
	}
\vspace{-3mm}
	\caption{
		Semantic compression performance on Action Recognition, MOT and VOS tasks.
		The plot titles are in \{Dataset\}-\{Model\} format.
	}
\vspace{-3mm}
	\label{fig:lbuv_mot_blind}
\end{figure*}

\begin{table*}[!t]
%	\vspace{-2mm}
	\caption{
		Results on action recognition in terms of Top1 accuracy (\%).
		`Upper-bound' is obtained by evaluating action recognition models with the original dataset.
		`K400' denotes Kinetics400.
	}
	\label{tab:benchmark_ar}
	\centering
	\tabcolsep=0.8mm
	\vspace{-2mm}
	\scalebox{0.85}{
		\begin{tabular}{|c|c|c|c|Hc|c|c|c|Hc|c|}
			\hline
			{}&\multicolumn{6}{c|}{\textbf{TSM}}  &\multicolumn{3}{c|}{\textbf{Slowfast}}&\multicolumn{2}{c|}{\textbf{TimeSformer}} \\
			\hline
			&HMDB51& Diving48	& UCF101 & HMDB51 & K400  & Diving48~ & UCF101 & HMDB51 & HMDB51 & UCF101 & HMDB51    \\
			\hline
			\textit{Bpp}& \textit{@0.02}& \textit{@ 0.05}&\textit{@0.04}  & \textit{@0.04} &  \textit{@0.06}  & \textit{@ 0.07~} &\textit{@0.03} & \textit{@0.02} & \textit{@0.03} &  \textit{@0.04} & \textit{@0.04}   \\
			\arrayrulecolor{gray}\cdashline{1-12}[5pt/3pt]
			HEVC &33.66&  22.48&  52.70 & 36.64 &  35.28  & 37.24 &85.13 &47.78 &62.68 & 69.92 &37.45   \\
			
			\hline
			FVC~\cite{hu2021fvc}&32.48 & 22.94 & 64.61 &  47.54 &  37.23 &46.29 &79.43 & 52.15 & 59.81 & 72.47 & 45.11   \\
			\hline
			PLVC~\cite{yang2021perceptual}&43.92& 30.13 & 71.40 &  48.67 &  48.65  & 42.63  & 87.38 & 56.33 & 64.50 & 70.06 & 44.59\\
			\hline
			VVC & 34.35& 42.75 & 76.97 & 55.72 & 49.11 & 53.23  &86.93 & 59.95 & 65.74 & 85.10  & 57.83\\
			\hline
			ROI~\cite{cai2021novel}&35.81 &44.94&78.63&56.33&49.82&55.64&87.25&60.44&66.62&85.91&58.21 \\
			\hline
			JPD-SE~\cite{duan2022jpd} &38.76&50.27&78.85&55.48&51.02 &57.22&84.86&60.22&64.76&84.03&57.79 \\
			\hline
			DCVC-DC~\cite{li2023neural} & 35.26&  {44.07}& {85.39} &61.40&  {55.81}  &{55.93}  &{86.27} &{60.83}  &{65.44} &{87.66}   &{55.34}\\
			\hline
			SMC~\cite{tian2023non}  & 52.12&{51.36}& {86.46}&  {62.92} &  {59.26}   &{60.16} &{88.89} &{63.48}  &{67.30} &{89.60}   &{60.74}\\
			\hline
			Ours& \textbf{62.24}&\textbf{63.37} & \textbf{89.16} &  \textbf{65.72} &  \textbf{61.83}   &\textbf{67.83} &\textbf{90.53} &\textbf{65.42}  &\textbf{68.84} &\textbf{92.16}   &\textbf{64.87}\\
			\arrayrulecolor{gray}\cdashline{1-12}[5pt/3pt]
			\textcolor{gray}{Upper-bound}&  \textcolor{gray}{72.81}& \textcolor{gray}{75.99} & \textcolor{gray}{93.97}  &  \textcolor{gray}{72.81} &  \textcolor{gray}{70.73} & \textcolor{gray}{75.99}
			& \textcolor{gray}{94.92} &  \textcolor{gray}{72.03} &  \textcolor{gray}{72.03} & \textcolor{gray}{95.43} &  \textcolor{gray}{71.44}
			\\
			\hline
			
		\end{tabular}
	}
\end{table*}

\begin{table}[!thbp]
%	\vspace{-2mm}
	\centering
	\caption{
		MOT and VOS performances of different coding methods.
		\greendownarrow ~denotes the lower is better.
		`Upper-bound' is obtained by evaluating the task models with the original datasets.
	}
	\vspace{-2mm}
	\label{tab:VOS_MOT_SOTA}
	\tabcolsep=1mm
	\scalebox{0.85}{
		\begin{tabular}{|c|c|c|c|c|c|c|c|c|}
			\hline
			&\multicolumn{4}{|c|}{MOT: ByteTrack on MOT17@0.01bpp}&\multicolumn{4}{c|}{VOS: XMEM on DAVIS2017@0.01bpp	}\\
			\hline
			&{\makecell{MOTA (\%)\greenuparrow}} & {\makecell{MOTP (\%)\greendownarrow}}
			& {\makecell{IDF1 (\%)\greenuparrow}}& {\makecell{FN\greendownarrow}}&{$\mathcal{J}$\&$\mathcal{F}$} (\%) & \textbf{$\mathcal{J}$} (\%) &\textbf{$\mathcal{F}$} (\%) & {$\mathcal{F}$-$Recal$} (\%)    \\
			\hline
			{{HEVC}}&61.30  &19.88 &64.32 &17377&57.68  &56.84 &58.51 &67.44\\
			\hline
			{{FVC~\cite{hu2021fvc}}}  &44.24  &21.97&52.53 &27508 &62.39  &61.22 &63.55 &75.67 \\
			\hline
			{{PLVC~\cite{yang2021perceptual}}}  &67.87  &18.44&68.95 &13299&61.45  &60.02 &62.87 &74.07 \\
			\hline
			{{VVC}}  &64.86  &19.49 &68.99 &15621    &67.47  &65.59 &69.36 &80.92 \\
			\hline
			ROI~\cite{cai2021novel} &65.29  &19.31 &67.78 &15270 &69.22  &67.16 &71.28 &83.80  \\
			\hline
			JPD-SE~\cite{duan2022jpd}&60.05  &20.62 &63.52 &17847  &61.48  &59.86 &63.09 &73.31  \\
			\hline
			DCVC-DC~\cite{li2023neural}&65.22  &18.22 &70.51 &12328  &72.86  &69.26 &76.46 &88.17  \\
			\hline
			{{SMC~\cite{tian2023non}}} &{70.84}  &{17.79} &{71.89} &{11710} &{74.20}  &{70.21} &{78.19} &{91.10} \\
			\hline
			Ours &\textbf{72.80}  &\textbf{16.61} &\textbf{73.21} &\textbf{10445}  &\textbf{76.85}  &\textbf{72.70} &\textbf{80.99} &\textbf{92.77} \\
			\arrayrulecolor{gray}\cdashline{1-9}[5pt/3pt]
			%			\hline
			{\textcolor{gray}{Upper-bound}} &\textcolor{gray}{78.60}  &\textcolor{gray}{15.80} &\textcolor{gray}{79.00} &\textcolor{gray}{7000} &\textcolor{gray}{87.70}  &\textcolor{gray}{84.06} &\textcolor{gray}{91.33} &\textcolor{gray}{97.02} \\
			\hline
		\end{tabular}
	}
	\vspace{-2mm}
\end{table}

From the above comparison, we observe that an advanced video semantic compression system requires the particular preservation of semantics, which is lacking in plain codecs like VVC and DCVC-DC. Furthermore, semantics derived from VFMs are more effective than SSL-learned semantics in SMC and hand-crafted methods like ROI. Additionally, our approach shows remarkable improvements on the challenging Diving48 dataset, which requires reasoning about fine-grained temporal relationships. The significant progress can be attributed to the proposed inter-frame entropy model, which compresses video semantics by predicting temporal dynamics.
Our approach is also effective on more advanced action networks like Slowfast (+1.94\% on HMDB51@0.02bpp) and TimeSformer (+2.56\% on UCF101@0.04bpp).
%We also illustrate rate-performance (RP) curves in the top row of Figure~\ref{fig:lbuv_mot_blind}.
% show our method outperforming all others by a large margin on all bitrate settings. 
%For curves of Slowfast and TimeSformer models, please refer to the supplementary material.

Finally, we also compare different coding approaches on the fine-grained action recognition dataset FineGym99~\cite{shao2020finegym}. Our approach (71.51\% Top1 accuracy) substantially outperforms VVC (62.43\%) and SMC (42.84\%) by a larger margin at 0.02bpp bitrate level, when using TSM as the recognition network.

\textbf{Multiple Object Tracking (MOT).}
In addition to the basic action recognition task, we also compare the coding methods on a much more challenging MOT task.
%This task requires not only inducing the local location of each object, but also extracting occlusion-robust appearance features for these objects.
As shown in the left part of Table~\ref{tab:VOS_MOT_SOTA}, our method achieves the state-of-the-art performance in terms of all metrics.
For example, our approach outperforms DCVC-DC and SMC by 7.58\% and 1.96\%  on the MOT17@0.01bpp setting, respectively, in terms of MOTA.
Finally, we provide the RP curves on MOT task in the second row of Figure~\ref{fig:lbuv_mot_blind}.
%It is observed that our method consistently outperforms other ones across different bitrate levels.

\textbf{Video Object Segmentation (VOS).}
We further compare our method with previous ones on VOS task, which requires more fine-grained semantic features than the above two tasks.
As shown in the right part of Table~\ref{tab:VOS_MOT_SOTA}, our method achieves the best performance in terms of all metrics.
For example, our approach outperforms DCVC-DC and SMC by 3.99\% and 2.65\% on the DAVIS2017@0.01bpp setting, respectively, in terms of $\mathcal{J}$\&$\mathcal{F}$.
We also illustrate the RP curves on VOS task in the third row of Figure~\ref{fig:lbuv_mot_blind}.
% our framework consistently outperforms other methods.

\begin{figure}[!t]
	%	\vspace{-3mm}
	\tabcolsep=0.5mm
	\centering
	\scriptsize
	\newcommand{\widmy}{0.19}
	\begin{tabular}{ccccc} 
		\includegraphics[width=\widmy \linewidth ]{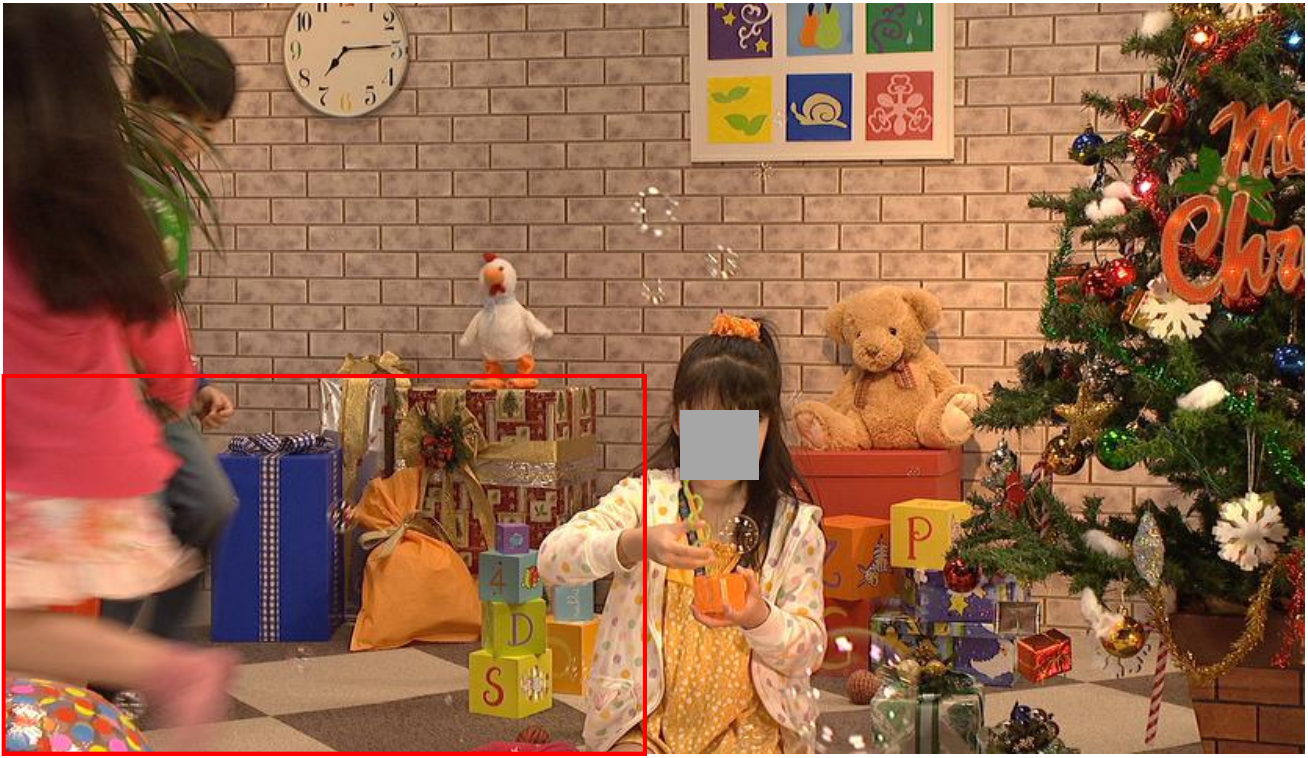}&
		\includegraphics[width=\widmy\linewidth]{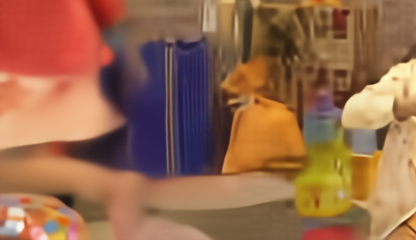}&
		\includegraphics[width=\widmy \linewidth]{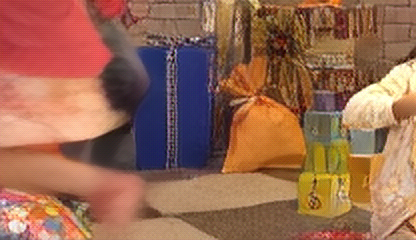}&
		\includegraphics[width=\widmy \linewidth]{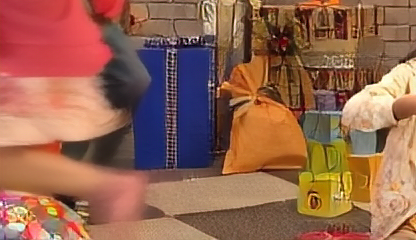}&
		\includegraphics[width=\widmy \linewidth]{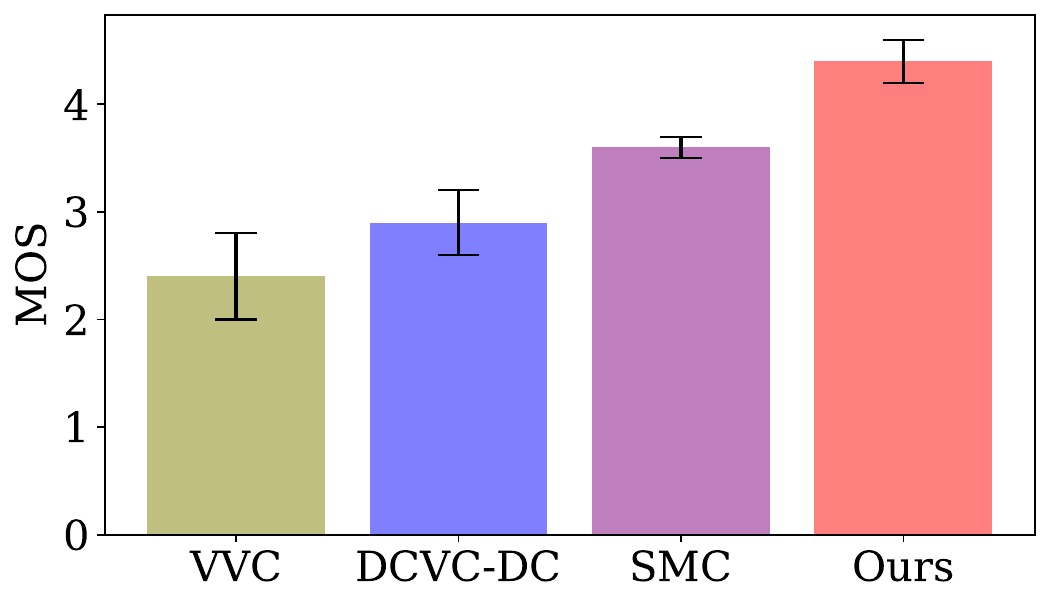}\\
		(a) Original & (b) DCVC-DC (1371$\times$) &
		(c) SMC (1403$\times$) & (d) Ours (1519$\times$) & (e) MOS comparison
	\end{tabular}
	\vspace{-2mm}
	\caption{
		Qualitative comparison between the compressed frame by different methods. The frame is from HEVC Class C dataset.
		The numbers in parentheses indicates the compression ratio.
		%		We mask the face region to avoid the potential privacy problem.
	}
	\vspace{-7mm}
	\label{fig:qua_results}
\end{figure}

\textbf{Visual Quality.}
As shown in Figure~\ref{fig:qua_results}, when compressing the videos by over 1300 times, our method shows better visual quality than DCVC-DC and SMC, \textit{e.g.}, sharper edges or less checkerboard artifacts, while with larger compression ratio (1519$\times$).
We also recruit four volunteers to score semantic completeness of the compressed frames from HEVC Class C dataset~\cite{sullivan2012overview} by different coding methods.
% which are compressed by different methods at the 0.02 bpp level.
The score range is from one to five, with higher scores indicating that objects and events within videos are more easily identifiable.
As shown in Figure~\ref{fig:qua_results} (e), our method achieves remarkably higher mean opinion score (MOS)~\cite{huynh2010study}.
Furthermore, our method achieves lower LPIPS~\cite{zhang2018unreasonable} than SMC by about 0.05 on average across all bitrate levels.
%When measured by PSNR metric, our method is inferior to the PSNR-oriented codec DCVC-DC on HEVC Class C,
% which is reasonable from the perceptual-distortion trade-off theory~\cite{blau2018perception}\cite{blau2019rethinking}.

%Compared with images, encoding settings for videos are much various.
\vspace{-4mm}
\subsection{Ablation Study and Analysis}\label{sec:method_ana}
\vspace{-2mm}
To validate each design of the proposed framework, we perform thorough ablation studies. For simplicity, numerical results in tables and text are reported based on the UCF101@0.02bpp setting using the TSM action network, unless stated otherwise.

\begin{figure}[!thbp]
		\renewcommand\arraystretch{0.10}
		\vspace{-6mm}
	\tabcolsep=1mm
	\centering
	\begin{tabular}{ccc}  
		\includegraphics[width=0.30\textwidth]{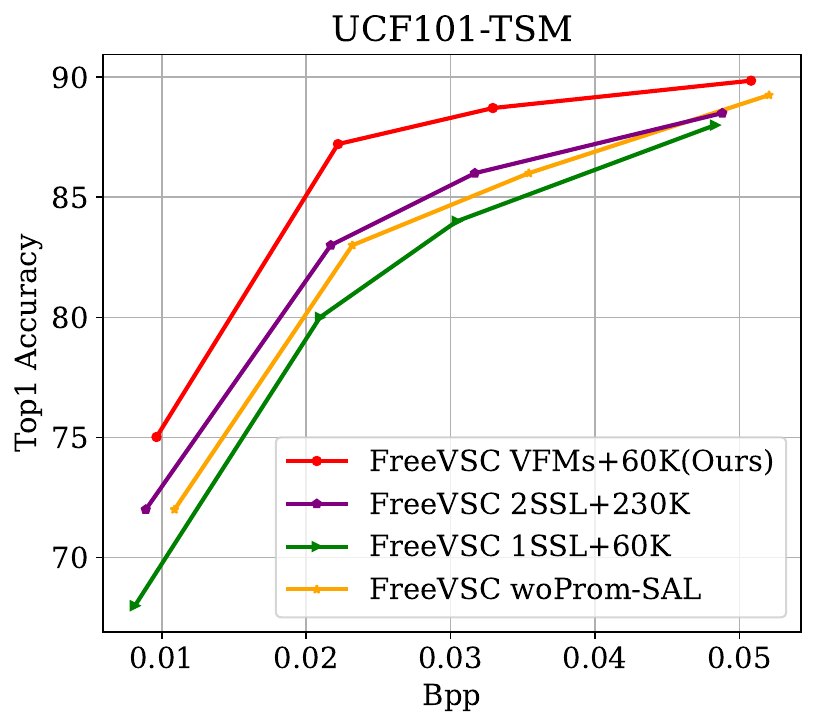}&
		\includegraphics[width=0.30 \textwidth]{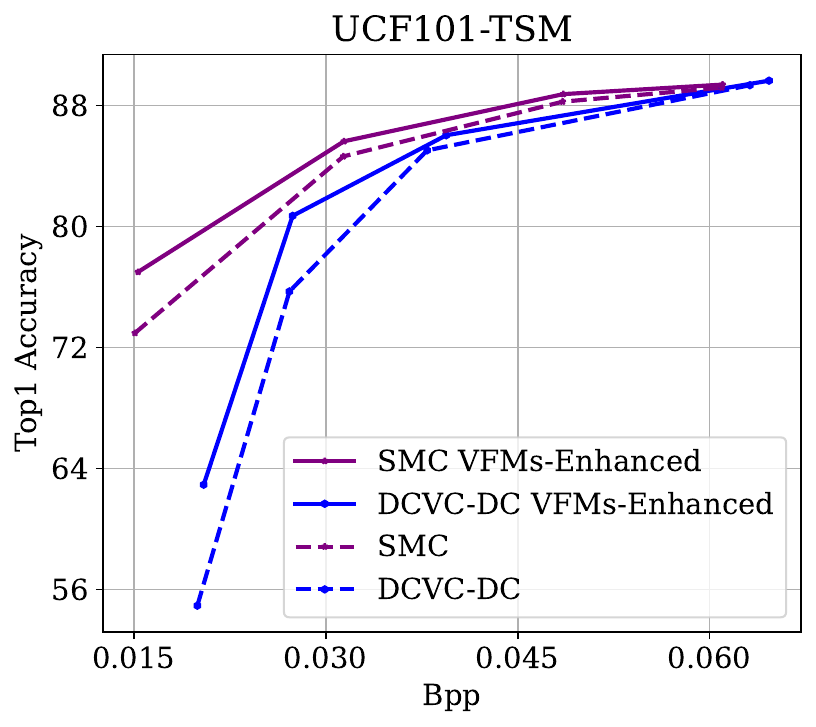}&
		\includegraphics[width=0.30\textwidth]{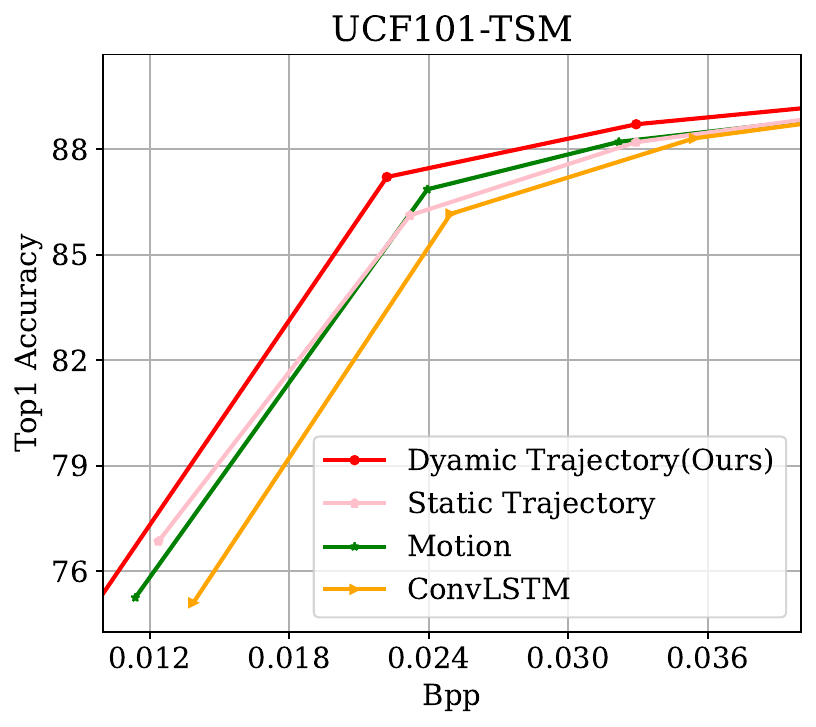}\\
		(a)&(b)&(c) 
	\end{tabular}
	\vspace{-4mm}
	\caption{
		(a) Ablation on the framework.
		(b) Effectiveness of introducing VFMs semantics to other approaches.
		(c) Comparison of different entropy models.
		`wo' denotes without this component.
	}
	\label{tab:ablation_framework}
	\vspace{-6mm}
\end{figure}

\textbf{Effectiveness of Each Components.}
As depicted in Figure~\ref{tab:ablation_framework} (a), to demonstrate the superiority of rich semantics from VFMs, we first replace the VFMs-based objective with a self-supervised learning (SSL) objective, specifically the same MAE loss in SMC~\cite{tian2023non}, training the model with the same 60K training videos as ours. The resulting FreeVSC 1SSL+60K model can be considered an enhanced version of SMC~\cite{tian2023non}, incorporating our network architecture and employing the same learning objective as SMC, thus enabling a fair comparison of different semantic learning objectives. The FreeVSC 1SSL+60K model performs much worse than our model, exhibiting over a 5\% accuracy drop at 0.02bpp.
This proves that a single SSL objective learns much weaker semantic representations than pre-trained VFMs.

To investigate whether a more diverse SSL objective and training data are beneficial, we train the model with a combined objective~\cite{huang2023contrastive} of MAE and contrastive learning, along with all 230K videos from Kinetics. The resulting FreeVSC 2SSL+230K model indeed improves performance over FreeVSC 1SSL+60K by approximately 2\% due to learning richer semantics, but remains inferior to our model by about 2\%. This underscores that previous unsupervised methods, such as SMC, are indeed constrained by their limited learning objective and data scale.
In contrast, our approach directly absorbs and inherits the rich semantics from the free off-the-shelf VFMs, which are learned with more diverse objectives and more video data, resulting in stronger results.
%The inherited rich semantics contributes to strong results with limited training data.

Furthermore, we apply the proposed VFMs-based semantic learning scheme directly to DCVC-DC and SMC. As depicted in Figure~\ref{tab:ablation_framework} (b), we observe significant performance improvements in both methods, affirming the generalizability of our approach. The rich semantics from off-the-shelf VFMs proves consistently beneficial for addressing the semantic compression problem, regardless of the specific network architectures of the compression network.

Subsequently, we investigate the neccestity of the Prom-SAL module. Concretely, we substitute the Prom-SAL module with a direct alignment scheme, employing simple linear layers to align the dimension of the compressed video semantics with that of VFM features. The resulting FreeVSC woProm-SAL model exhibits a severe performance drop of 5\% (see orange and red curves in Figure~\ref{tab:ablation_framework} (a)), attributed to the incapability of linear layers to effectively align the complex features from large-scale VFMs.

Finally, we demonstrate the superiority of our dynamic trajectory-based entropy coding scheme.
We first replace the video-adaptive dynamic trajectory design with the static trajectory that is fixed and shared by all videos.
As illustrated in Figure~\ref{tab:ablation_framework} (c), the performance degrades severely, since the fixed static trajectory can not well model the diverse contents of different videos.
Then, we compare our approach to previous popular inter-frame coding scheme, namely the motion compensation-based~\cite{hu2021fvc} and ConvLSTM-based~\cite{yang2020learning} schemes.
%The former focuses on modeling explicit object displacements, while the latter implicitly models video dynamics with hidden states.
%As illustrated in Figure~\ref{tab:ablation_framework} (c), 
It can be observed that both two models are inferior to ours, with approximately 1.5\% and 2.2\% accuracy drops for motion and ConvLSTM, respectively. The reasons may be that (1) explicit motion cannot effectively model high-level and abstract semantic dynamics, and (2) ConvLSTM struggles to adapt to video content. In contrast, our entropy model predicts customized non-linear transforms for each video, flexibly handling high-level information and adapting to video content.

\textbf{Advantage of Learning from Multiple VFMs.}
Table~\ref{tab:ablation_VFM} (a) presents the comparison of using different VFM. First, when employing a single VFM, the resulting models $M_a$ (VideoMAEv2) and $M_b$ (DINOv2) achieve unsatisfactory performances, with 83.11\% and 83.56\% Top1 accuracy, respectively. Furthermore, after leveraging both two above VFMs, the resulting model $M_c$ attains better result 84.11\% than using any single one of them, confirming that VFMs mutually enhance each other in our framework. Finally, scaling the number of VFMs by incorporating the SSL-Swin model~\cite{xie2021self}, the final model $M_{our}$ achieves the best result 85.07\%.

\begin{table}[!b]
	\vspace{-4mm}
	\caption{
		(a) Comparison of using different VFMs.
%		$M_b$ denotes the model trained by using a SSL loss, \textit{i.e.}, VideoMAE~\cite{tong2022videomae}.
%		Other models are trained by absorbing knowledge from the pre-trained VFMs.
%		For example, $M_e$ is trained with pre-trained VideoMAEv2 and DINOv models.
%		Using more VFMs consistently benefits our framework.
		(b) Visualization of compressed video features.
%		  by these models at the same bitrate level.
	}
	\vspace{-2mm}
	\label{tab:ablation_VFM}
	\begin{minipage}{0.44 \linewidth}
		
		\centering
		\tabcolsep=1mm
		\scalebox{0.9}{
			\begin{tabular}{|cH|c|c|c|cH|}
				
				\hline
				$Models$	&  $M_b$ & $M_a$&  $M_b$&  $M_c$&  $M_{our}$ &  $M_{f}$\\
				\hline
				{ VideoMAEv2} &\xmark&\cmark&\xmark&\cmark&\cmark&\cmark \\
				\hline
				{ DINOv2}     &\xmark&\xmark&\cmark&\cmark&\cmark&\cmark \\
				\hline
				{ SSL-Swin}   &\xmark&\xmark&\xmark&\xmark&\cmark&\cmark \\
				%			\hline
				%			Prom-SAL                        &\xmark&\xmark&\xmark&\cmark&\cmark&\xmark \\
				\hline
				Accuracy (\%)   & 	79.33	&83.11		& 83.56	       & 84.11    & \textbf{85.07}& {86.40} \\
				\hline
				\multicolumn{7}{c}{(a)}
			\end{tabular}
		}
	\end{minipage}
	\begin{minipage}{0.55 \linewidth}
			\renewcommand\arraystretch{0.90}
		\tabcolsep=0mm
		\centering
			\scalebox{0.9}{
		\begin{tabular}{cccc}  
			\includegraphics[width=0.24\textwidth]{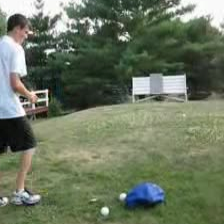}&\hspace{0.5mm}
			\includegraphics[width=0.24\textwidth]{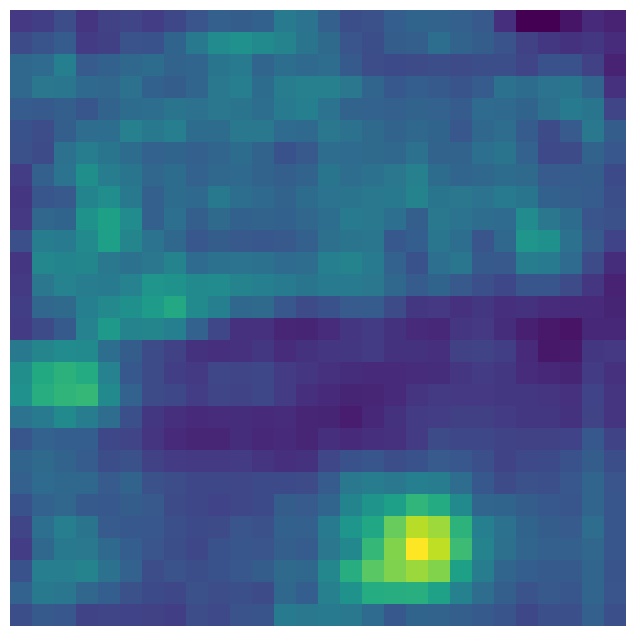}&
			\includegraphics[width=0.24 \textwidth]{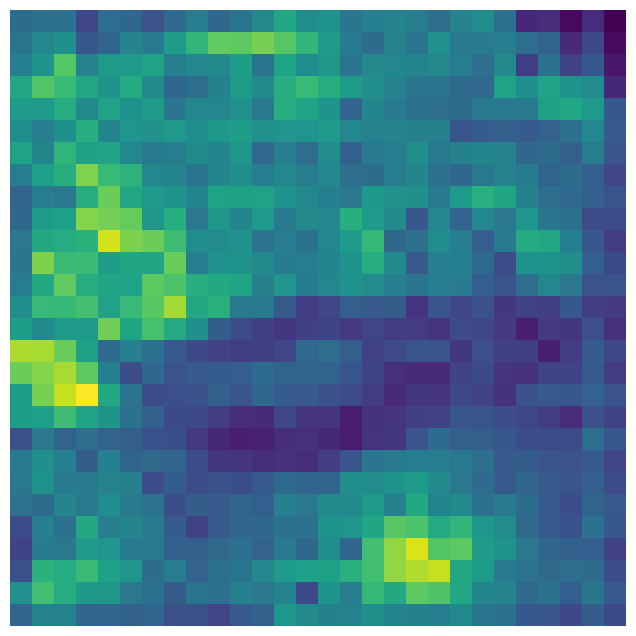}&
			\includegraphics[width=0.24\textwidth]{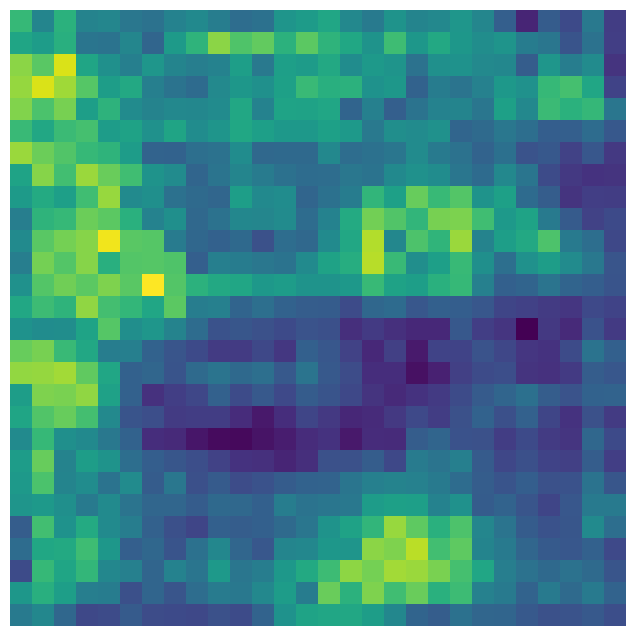}\\
			Frame & SSL & VideoMAEv2 & 2VFMs \\
			\multicolumn{4}{c}{(b)}
		\end{tabular}
	}
	\end{minipage}
	
\end{table}

Moreover, we visualize the feature maps of different models.
The features are output from the third stage of the decoder network at the 0.02bpp level.
As illustrated in Table~\ref{tab:ablation_VFM} (b), the model trained with the SSL objective in SMC~\cite{tian2023non} exhibits weakest semantic representation capability, where the `person' cannot be discriminated from the background effectively.
In contrast, the model trained with VFM VideoMAEv2 demonstrates strong semantics in temporally moving regions, such as `person arm and leg', while missing the static region `person head'.
Further, the model equipped with 2VFMs VideoMAEv2+DINOv2 preserves semantics more comprehensively, demonstrating that our framework absorbs complementary semantics from diverse VFMs.

\textbf{Ablation on Prom-SAL.}
As mentioned in Section~\ref{sec:method}, we introduce Prom-SAL, \textit{i.e.}, a VFM-shared transformation $\psi$ with VFM-specific prompts, to align the compressed video representation to features from different VFMs.
In this section, we conduct extensive experiments to verify the effectiveness of each design in Prom-SAL.

\begin{table}[!t]
%	\vspace{-4mm}
	\centering
	\caption{
		(a) Ablation Study on Prom-SAL. C and F denote coarse and fine-level features, respectively.
		(b) \textit{upper}: Comparison of different strategies for sharing the Prom-SAL parameters when aligning multiple VFMs.
		Gradient harmony is calculated as the cosine similarity between the gradients propagated from VideoMAEv2 and DINOv2.
		\textit{lower}: Comparison of different network architecture choices for the alignment layer $\psi$ in Prom-SAL.
	}
\vspace{-1mm}
\label{tab:ab_prom_sal}
	\begin{minipage}{0.49 \linewidth}
	\scalebox{0.9}{
		\centering
		\begin{tabular}{|cH|c|c|c|c|}
			\hline
			$Models$ & $M$ & $M_d$&  $M_e$&  $M_f$&  $M_{our}$\\
			\hline
			Shared-Transformer  &\xmark &\cmark&\cmark&\cmark&\cmark \\
			\hline
			\makecell[c]{Shared-Prompts}&\xmark &\cmark&\xmark&\xmark&\xmark \\
			\hline
			Feature Level         &C+F	&C+F&C-Only&F-Only&C+F \\
			\hline
			Accuracy (\%) & 	83.01	&80.92		& 83.07	       & 84.28    & \textbf{85.07} \\
			\hline
			\multicolumn{6}{c}{(a)}
		\end{tabular}
	}
	\end{minipage}
	\begin{minipage}{0.50 \linewidth}
		\vspace{-1mm}
	\begin{minipage}{1 \linewidth}
		\renewcommand\arraystretch{0.95}
		\scalebox{0.90}{
			\centering
			\begin{tabular}{|c|c|c|c|c|}
				%			\hline
				%			&\multicolumn{3}{c|}{Aligning to Three VFMs}\\
				\hline
				{ Parameter Share} &None &\makecell[c]{Adapter} &\makecell[c]{LORA} & \makecell[c]{Prompts} \\
				\hline
				Accuracy (\%)&83.01  &84.23 &84.97& \textbf{85.07} \\
				\hline
				 Gradient Harmony&0.42 &0.62 &0.87&\textbf{0.89} \\
				\hline
%				\multicolumn{5}{c}{(b)}
			\end{tabular}
		}
	\end{minipage}	

	\begin{minipage}{1 \linewidth}
		\renewcommand\arraystretch{0.95}
		\scalebox{0.90}{
		\centering
		\begin{tabular}{|c|c|c|c|}
			\hline
			{$\psi$ architecture}  &3D CNN &ST-divided Global& Vanilla \\
			\hline
			Accuracy (\%)&79.56 &82.76&\textbf{85.07} \\
%			\hline
%			Gradient Harmony&0.13 &0.69&\textbf{0.91} \\
			\hline
			\multicolumn{4}{c}{(b)}
		\end{tabular}
	}
	\end{minipage}	
	\end{minipage}
\vspace{-4mm}
\end{table}

As shown in Table~\ref{tab:ab_prom_sal} (a), after removing VFM-specific prompts, the resulting model $M_d$ learns average features of different VFMs.
The distorted average semantics leads to a 4.15\% accuracy drop. We then compare the impact of using different-level features. Using only coarse- or fine-level features yields unsatisfactory results, while our combination strategy achieves the best result, consistent with previous studies~\cite{lin2017feature}.

Further, we compare different strategies for sharing parameters of Prom-SAL across different VFMs.
As depicted in Table~\ref{tab:ab_prom_sal} (b) \textit{upper}, the absence of a parameter-sharing design results in a performance drop of over 2\%, attributed to gradient contradictions among different VFMs, i.e., low gradient harmony.
We then explore two recent strategies: Adapter~\cite{pan2022st} and LORA~\cite{2021LoRA}. Adapter involves inserting VFM-specific ST-Adapter blocks~\cite{pan2022st} into the shared Transformer, while LORA utilizes a low-rank adaption technique~\cite{2021LoRA}, introducing VFM-specific rank-4 matrices to tune attention weights of the shared Transformer.
ST-Adapter achieves the most inferior result, attributed to the inflexibility of the depth-3D convolution used. In contrast, more flexible methods like LORA and Prompts, initially proposed in the natural language processing (NLP) community, exhibit promising results, alongside high gradient harmony values. It is anticipated that leveraging more advanced parameter-sharing strategies from the NLP community will further enhance our framework.

Finally, we explore network architecture choices for Prom-SAL beyond the vanilla Transformer.
As depicted in Table~\ref{tab:ab_prom_sal} (b) \textit{lower}, employing 3D CNN or simplified spatial-temporal divided attentions~\cite{bertasius2021space} results in the accuracy drop of 5.51\% and 2.31\%, respectively. This proves the necessity of utilizing a flexible vanilla Transformer architecture for aligning the rich semantics from complex VFMs.

\textbf{Ablation on Dynamic Trajectory-based Compression Scheme.}
As shown in Table~\ref{tab:ablation_entropy} (a), after removing both the content-adaptive trajectory and region-wise trajectory designs, the resulting model $M_g$ increases the bitcost by 21.7\% compared to our model. Then, using a region-wise but static trajectory design leads to model $M_h$, which sees an 11.2\% increase in bitcost. Conversely, employing a content-adaptive yet region-shared trajectory design leads to model $M_i$, causing a 13.8\% increase in bitcost. Additionally, comparing models $M_j$ and $M_{our}$ reveals the benefits of long-term information in semantic compression.
Furthermore, investigating the impact of trajectory number $S$, we observe BD-rates of 21.7\%, 6.4\%, 0\%, and -1.8\% for $S=1$, 3, 5, and 10, respectively. We select $S=5$ to strike the best computational cost-performance trade-off.

Further, we visualize the predicted future frame features.
% by connecting it to the frame decoder using a single linear layer.
As shown in Table~\ref{tab:ablation_entropy} (b), the predicted future frame (the third figure), obtained from the two past historical frame features (the first two figures), accurately reflects the future semantics `the man bent down after throwing' of the ground-truth frame (the fourth figure).

% Encoding $\hat{f}_{4}$ with $z_4$ as the condition, which is already similar to $\hat{f}_{4}$, significantly reduces the bitcost by 4 times (0.017 vs. 0.073 bpp), compared to not using $z_4$.

\begin{table}[!t]
	\caption{
		(a) Ablation Study on Entropy Model.
		BD-Rate indicates the increased bitrates compared to our model. \textit{Lower is better.}
		(b) Visualization of the predicted future semantic feature.
%		 of current frame $z_4$ from historical feature $\hat{f}_3$, which is semantically similar to ground-truth feature $\hat{f}_4$.
%		 of the current frame.
%		We visualize the features by decoding them with the frame decoder network.
	}
\vspace{-2mm}
	\centering
	\begin{minipage}{0.51 \linewidth}
	\tabcolsep=1.15mm
	\renewcommand\arraystretch{0.95}
	\scalebox{0.9}{
		\begin{tabular}{|c|c|c|c|c|c|c|}
			\hline
				 & $M_g$&  $M_h$&$M_i$& $M_j$& $M_{our}$\\
			\hline
			Content-adaptive   &\xmark  &\xmark &\cmark&  \cmark&\cmark \\
			\hline
			Region-wise    &\xmark &\cmark  &\xmark&\cmark&\cmark \\
			\hline
			Long-term Fusion   &\cmark &\cmark  &\cmark&\xmark&\cmark  \\
%			Motion-compensation &\cmark &\xmark &\xmark&\xmark&\xmark  \\
			\hline
			BD-Rate (\%)  & 	21.7&11.2	&13.8		& 7.6	& \textbf{0} \\
			\hline
			\multicolumn{6}{c}{(a)}
		\end{tabular}
	}
	\end{minipage}
	\begin{minipage}{0.48 \linewidth}
			\renewcommand\arraystretch{0.80}
			\newcommand{\widmy}{0.31}
		\tabcolsep=0.35mm
			\scalebox{0.90}{
		\begin{tabular}{cccc}
			\includegraphics[width=0.26 \textwidth]{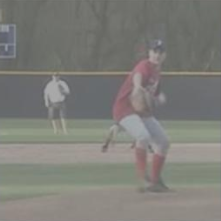}&
		\includegraphics[width=0.259 \textwidth]{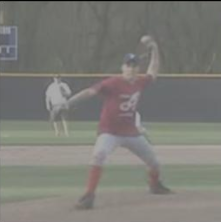}&
		\includegraphics[width=0.264 \textwidth] {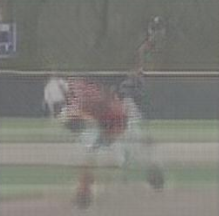}&
		\includegraphics[width=0.266 \textwidth]{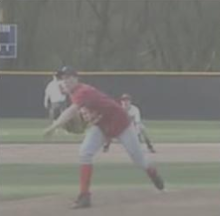}
		\\
		\multicolumn{2}{c}{Historical two frames} & Predicted  & Ground-truth \\
		\multicolumn{4}{c}{(b)}
	\end{tabular}
}
	\end{minipage}
	\label{tab:ablation_entropy}
	\vspace{-6mm}
\end{table}

%
%\begin{figure}[!thbp]
%
%	\centering
%	\
%	\vspace{-3mm}
%	$\hat{f}_{1}$ &  $\hat{f}_{2}$ & 
%	\caption{
%		Visualization of the predicted context.
%		We visualize the features by decoding them with the frame decoder network.
%	}
%	\label{fig:vis_trajectory}
%\end{figure}

%To compare our scheme with previous motion-based ones, we employ the MEMC (motion estimation and motion compensation) networks in FVC~\cite{hu2021fvc} to compress the semantic features.
%To enable fair comparison, this baseline is also fused with the long-term historical information and with similar parameter number.
%The resulted $M_j$ leads to 15.6\% bitrate increase,
%proving that our trajectory prediction-based scheme is more effective for semantic compression.

%\textbf{Model Complexity.}

\textbf{Model Complexity.}
The parameter numbers of our encoder and decoder networks are 10.4M and 9.3M, respectively.
We report the per frame running time of a 1080p video on the machine with a Nvidia 2080Ti GPU.
Our encoding time is 722ms, which is better than DCVC-DC (1005ms) and SMC (1413ms).
Our decoding time is 226ms, which is 3$\times$ faster than DCVC-DC (765ms).
While our model is trained with several large VFMs, they are only utilized during training. Therefore, our model still maintains a fast inference speed, since most layers of our networks are operated on low-resolution feature maps ($\frac{1}{8}$, $\frac{1}{16}$ and $\frac{1}{32}$ resolution scales).

\vspace{-4mm}
\section{Conclusion, Limitation, and Potential Negative Impact}
\vspace{-4mm}
\textbf{Conclusion.} In this paper, we have introduced Free-VSC, the first video semantic compression framework leveraging VFMs. To enable VFMs to collaboratively guide the learning of the compression model, a prompt-based semantic alignment layer is introduced. Additionally, we have proposed a trajectory-based dynamic entropy model for more efficient inter-frame semantic compression. Comprehensive experiments validate our approach's superiority in three video analysis tasks.

\textbf{Limitation and Future Work.}
%We note that the proposed Free-VSC is still a preliminary effort for leveraging VFMs to enhance video semantic compression.
%Certain limitations can be observed, which are also interesting future research directions.
First, the number of VFMs is rapidly increasing. It is possible that some of them are not helpful to the semantic compression task. A fast method for judging their effectiveness is missed in this work and will be developed in the future.
Second, same VFMs are employed across different bitrates in this work for simplicity. However, different VFMs usually contain knowledge of different granularity.
A bitrate-adaptive VFM selection strategy will be devised in the future.
Besides, some hyper-parameters in this work rely on empirical selection, which can be automatically decided by the reinforce learning algorithms.

\textbf{Potential Negative Impact.}
The proposed Free-VSC extends the boundaries of the minimum bitcost required to preserve general semantic information within videos. The concept and technique are general and unbiased toward any specific video task, as we avoid using task-specific annotations.
However, despite its technical contributions and our original intention not to favor any particular application, concerns remain regarding its potential illegal deployment in applications that may violate privacy rights and human rights, such as video surveillance and behavior monitoring.

\noindent \textbf{Acknowledgment.} This work is supported by National Natural Science Foundation of China (62225112), Shanghai Artificial Intelligence Laboratory, and National Natural Science Foundation of China under Grant 62101326.

%Another 
%Future work will focus on adaptively controlling transferring strategy for different bitrate settings.
%There limitations will be left for the interesting future researches.

%\section{}
\bibliographystyle{splncs04}
\bibliography{main}

\begin{thebibliography}{100}
\providecommand{\url}[1]{\texttt{#1}}
\providecommand{\urlprefix}{URL }
\providecommand{\doi}[1]{https://doi.org/#1}

\bibitem{mmtrack2020}
Mmtracking: Openmmlab video perception toolbox and benchmark.
  \url{https://github.com/open-mmlab/mmtracking} (2020)

\bibitem{2020mmaction2}
Openmmlab's next generation video understanding toolbox and benchmark.
  \url{https://github.com/open-mmlab/mmaction2} (2020)

\bibitem{akbari2019dsslic}
Akbari, M., Liang, J., Han, J.: Dsslic: Deep semantic segmentation-based
  layered image compression. In: International Conference on Acoustics, Speech
  and Signal Processing (2019)

\bibitem{bai2022towards}
Bai, Y., Yang, X., Liu, X., Jiang, J., Wang, Y., Ji, X., Gao, W.: Towards
  end-to-end image compression and analysis with transformers. In: Annual AAAI
  Conference on Artificial Intelligence (2022)

\bibitem{balle2018variational}
Ball{\'e}, J., Minnen, D., Singh, S., Hwang, S.J., Johnston, N.: Variational
  image compression with a scale hyperprior. In: International Conference on
  Learning Representations (2018)

\bibitem{bertasius2021space}
Bertasius, G., Wang, H., Torresani, L.: Is space-time attention all you need
  for video understanding? In: International Conference on Machine Learning
  (2021)

\bibitem{bross2021overview}
Bross, B., Wang, Y.K., Ye, Y., Liu, S., Chen, J., Sullivan, G.J., Ohm, J.R.:
  Overview of the versatile video coding (vvc) standard and its applications.
  IEEE Transactions on Circuits and Systems for Video Technology  (2021)

\bibitem{cai2021novel}
Cai, Q., Chen, Z., Wu, D.O., Liu, S., Li, X.: A novel video coding strategy in
  hevc for object detection. IEEE Transactions on Circuits and Systems for
  Video Technology  (2021)

\bibitem{cao2023observation}
Cao, J., Pang, J., Weng, X., Khirodkar, R., Kitani, K.: Observation-centric
  sort: Rethinking sort for robust multi-object tracking. In: Proceedings of
  the IEEE/CVF Conference on Computer Vision and Pattern Recognition (2023)

\bibitem{caron2020unsupervised}
Caron, M., Misra, I., Mairal, J., Goyal, P., Bojanowski, P., Joulin, A.:
  Unsupervised learning of visual features by contrasting cluster assignments.
  In: Advances in Neural Information Processing Systems (2020)

\bibitem{caron2021emerging}
Caron, M., Touvron, H., Misra, I., J{\'e}gou, H., Mairal, J., Bojanowski, P.,
  Joulin, A.: Emerging properties in self-supervised vision transformers. In:
  Proceedings of the IEEE/CVF International Conference on Computer Vision
  (2021)

\bibitem{carreira2017quo}
Carreira, J., Zisserman, A.: Quo vadis, action recognition? a new model and the
  kinetics dataset. In: Proceedings of the IEEE/CVF Conference on Computer
  Vision and Pattern Recognition (2017)

\bibitem{chao2015keypoint}
Chao, J., Steinbach, E.: Keypoint encoding for improved feature extraction from
  compressed video at low bitrates. IEEE Transactions on Multimedia  (2015)

\bibitem{che2021adversarial}
Che, Z., Borji, A., Zhai, G., Ling, S., Li, J., Tian, Y., Guo, G., Le~Callet,
  P.: Adversarial attack against deep saliency models powered by non-redundant
  priors. IEEE Transactions on Image Processing  (2021)

\bibitem{chen2024cross}
Chen, H., Qu, Z., Tian, Y., Jiang, N., Qin, Y., Gao, J., Zhang, R., Ma, Y.,
  Jin, Z., Zhai, G.: A cross-temporal multimodal fusion system based on deep
  learning for orthodontic monitoring. Computers in Biology and Medicine
  (2024)

\bibitem{chen2020improved}
Chen, X., Fan, H., Girshick, R., He, K.: Improved baselines with momentum
  contrastive learning. arXiv preprint arXiv:2003.04297  (2020)

\bibitem{chen2023transtic}
Chen, Y.H., Weng, Y.C., Kao, C.H., Chien, C., Chiu, W.C., Peng, W.H.: Transtic:
  Transferring transformer-based image compression from human perception to
  machine perception. In: Proceedings of the IEEE/CVF International Conference
  on Computer Vision (2023)

\bibitem{chen2020dynamic}
Chen, Y., Dai, X., Liu, M., Chen, D., Yuan, L., Liu, Z.: Dynamic convolution:
  Attention over convolution kernels. In: Proceedings of the IEEE/CVF
  Conference on Computer Vision and Pattern Recognition (2020)

\bibitem{chen2019lossy}
Chen, Z., Fan, K., Wang, S., Duan, L.Y., Lin, W., Kot, A.: Lossy intermediate
  deep learning feature compression and evaluation. In: ACM International
  Conference on Multimedia (2019)

\bibitem{chen2019toward}
Chen, Z., Fan, K., Wang, S., Duan, L., Lin, W., Kot, A.C.: Toward intelligent
  sensing: Intermediate deep feature compression. IEEE Transactions on Image
  Processing  (2019)

\bibitem{chen2024gaia}
Chen, Z., Sun, W., Tian, Y., Jia, J., Zhang, Z., Wang, J., Huang, R., Min, X.,
  Zhai, G., Zhang, W.: Gaia: Rethinking action quality assessment for
  ai-generated videos. arXiv preprint arXiv:2406.06087  (2024)

\bibitem{cheng2022xmem}
Cheng, H.K., Schwing, A.G.: Xmem: Long-term video object segmentation with an
  atkinson-shiffrin memory model. In: European Conference on Computer Vision.
  pp. 640--658. Springer (2022)

\bibitem{choi2018high}
Choi, H., Bajic, I.V.: High efficiency compression for object detection. In:
  International Conference on Acoustics, Speech and Signal Processing (2018)

\bibitem{choi2018near}
Choi, H., Baji{\'c}, I.V.: Near-lossless deep feature compression for
  collaborative intelligence. In: International Workshop on Multimedia Signal
  Processing (2018)

\bibitem{choi2022scalable11}
Choi, H., Baji{c}, I.V.: Scalable image coding for humans and machines. IEEE
  Transactions on Image Processing  (2022)

\bibitem{choi2020task}
Choi, J., Han, B.: Task-aware quantization network for jpeg image compression.
  In: European Conference on Computer Vision (2020)

\bibitem{dosovitskiy2020image}
Dosovitskiy, A., Beyer, L., Kolesnikov, A., Weissenborn, D., Zhai, X.,
  Unterthiner, T., Dehghani, M., Minderer, M., Heigold, G., Gelly, S., et~al.:
  An image is worth 16x16 words: Transformers for image recognition at scale.
  In: International Conference on Learning Representations (2020)

\bibitem{duan2022develop}
Duan, H., Shen, W., Min, X., Tian, Y., Jung, J.H., Yang, X., Zhai, G.: Develop
  then rival: A human vision-inspired framework for superimposed image
  decomposition. IEEE Transactions on Multimedia  (2022)

\bibitem{duan2015overview}
Duan, L.Y., Chandrasekhar, V., Chen, J., Lin, J., Wang, Z., Huang, T., Girod,
  B., Gao, W.: Overview of the mpeg-cdvs standard. IEEE Transactions on Image
  Processing  (2015)

\bibitem{duan2013compact}
Duan, L.Y., Gao, F., Chen, J., Lin, J., Huang, T.: Compact descriptors for
  mobile visual search and mpeg cdvs standardization. In: IEEE International
  Symposium on Circuits and Systems (2013)

\bibitem{duan2018compact}
Duan, L.Y., Lou, Y., Bai, Y., Huang, T., Gao, W., Chandrasekhar, V., Lin, J.,
  Wang, S., Kot, A.C.: Compact descriptors for video analysis: The emerging
  mpeg standard. IEEE Transactions on Multimedia  (2018)

\bibitem{duan2020video}
Duan, L., Liu, J., Yang, W., Huang, T., Gao, W.: Video coding for machines: A
  paradigm of collaborative compression and intelligent analytics. IEEE
  Transactions on Image Processing  (2020)

\bibitem{duan2022jpd}
Duan, S., Chen, H., Gu, J.: Jpd-se: High-level semantics for joint
  perception-distortion enhancement in image compression. IEEE Transactions on
  Image Processing  (2022)

\bibitem{dubois2021lossy}
Dubois, Y., Bloem-Reddy, B., Ullrich, K., Maddison, C.J.: Lossy compression for
  lossless prediction. In: Advances in Neural Information Processing Systems
  (2021)

\bibitem{esser2021taming}
Esser, P., Rombach, R., Ommer, B.: Taming transformers for high-resolution
  image synthesis. In: Proceedings of the IEEE/CVF Conference on Computer
  Vision and Pattern Recognition (2021)

\bibitem{fang2022prior}
Fang, Z., Shen, L., Li, M., Wang, Z., Jin, Y.: Prior-guided contrastive image
  compression for underwater machine vision. IEEE Transactions on Circuits and
  Systems for Video Technology  (2022)

\bibitem{feichtenhofer2019slowfast}
Feichtenhofer, C., Fan, H., Malik, J., He, K.: Slowfast networks for video
  recognition. In: Proceedings of the IEEE/CVF International Conference on
  Computer Vision (2019)

\bibitem{feng2022image}
Feng, R., Jin, X., Guo, Z., Feng, R., Gao, Y., He, T., Zhang, Z., Sun, S.,
  Chen, Z.: Image coding for machines with omnipotent feature learning. In:
  European Conference on Computer Vision (2022)

\bibitem{galteri2018video}
Galteri, L., Bertini, M., Seidenari, L., Del~Bimbo, A.: Video compression for
  object detection algorithms. In: International Conference on Pattern
  Recognition (2018)

\bibitem{ge2024task}
Ge, X., Luo, J., Zhang, X., Xu, T., Lu, G., He, D., Geng, J., Wang, Y., Zhang,
  J., Qin, H.: Task-aware encoder control for deep video compression. In:
  Proceedings of the IEEE/CVF Conference on Computer Vision and Pattern
  Recognition (2024)

\bibitem{goodfellow2020generative}
Goodfellow, I., Pouget-Abadie, J., Mirza, M., Xu, B., Warde-Farley, D., Ozair,
  S., Courville, A., Bengio, Y.: Generative adversarial networks.
  Communications of the ACM  (2020)

\bibitem{grill2020bootstrap}
Grill, J.B., Strub, F., Altch{\'e}, F., Tallec, C., Richemond, P., Buchatskaya,
  E., Doersch, C., Avila~Pires, B., Guo, Z., Gheshlaghi~Azar, M., et~al.:
  Bootstrap your own latent-a new approach to self-supervised learning (2020)

\bibitem{he2022masked}
He, K., Chen, X., Xie, S., Li, Y., Doll{\'a}r, P., Girshick, R.: Masked
  autoencoders are scalable vision learners. In: Proceedings of the IEEE/CVF
  Conference on Computer Vision and Pattern Recognition (2022)

\bibitem{he2020momentum}
He, K., Fan, H., Wu, Y., Xie, S., Girshick, R.: Momentum contrast for
  unsupervised visual representation learning. In: Proceedings of the IEEE/CVF
  Conference on Computer Vision and Pattern Recognition (2020)

\bibitem{2021LoRA}
Hu, E.J., Shen, Y., Wallis, P., Allen-Zhu, Z., Li, Y., Wang, S., Wang, L.,
  Chen, W.: Lora: Low-rank adaptation of large language models. arXiv preprint
  arXiv:2106.09685  (2021)

\bibitem{hu2020towards}
Hu, Y., Yang, S., Yang, W., Duan, L.Y., Liu, J.: Towards coding for human and
  machine vision: A scalable image coding approach. In: International
  Conference on Multimedia and Expo (2020)

\bibitem{hu2022coarse}
Hu, Z., Lu, G., Guo, J., Liu, S., Jiang, W., Xu, D.: Coarse-to-fine deep video
  coding with hyperprior-guided mode prediction. In: Proceedings of the
  IEEE/CVF Conference on Computer Vision and Pattern Recognition (2022)

\bibitem{hu2021fvc}
Hu, Z., Lu, G., Xu, D.: Fvc: A new framework towards deep video compression in
  feature space. In: Proceedings of the IEEE/CVF Conference on Computer Vision
  and Pattern Recognition (2021)

\bibitem{huang2023contrastive}
Huang, Z., Jin, X., Lu, C., Hou, Q., Cheng, M.M., Fu, D., Shen, X., Feng, J.:
  Contrastive masked autoencoders are stronger vision learners. IEEE
  Transactions on Pattern Analysis and Machine Intelligence  (2023)

\bibitem{huang2021visual}
Huang, Z., Jia, C., Wang, S., Ma, S.: Visual analysis motivated rate-distortion
  model for image coding. In: International Conference on Multimedia and Expo
  (2021)

\bibitem{huang2022hmfvc}
Huang, Z., Jia, C., Wang, S., Ma, S.: Hmfvc: A human-machine friendly video
  compression scheme. IEEE Transactions on Circuits and Systems for Video
  Technology  (2022)

\bibitem{huynh2010study}
Huynh-Thu, Q., Garcia, M.N., Speranza, F., Corriveau, P., Raake, A.: Study of
  rating scales for subjective quality assessment of high-definition video.
  IEEE Transactions on Broadcasting  (2010)

\bibitem{isola2017image}
Isola, P., Zhu, J.Y., Zhou, T., Efros, A.A.: Image-to-image translation with
  conditional adversarial networks. In: Proceedings of the IEEE/CVF Conference
  on Computer Vision and Pattern Recognition (2017)

\bibitem{jia2022visual}
Jia, M., Tang, L., Chen, B.C., Cardie, C., Belongie, S., Hariharan, B., Lim,
  S.N.: Visual prompt tuning. In: European Conference on Computer Vision (2022)

\bibitem{kasturi2008framework}
Kasturi, R., Goldgof, D., Soundararajan, P., Manohar, V., Garofolo, J., Bowers,
  R., Boonstra, M., Korzhova, V., Zhang, J.: Framework for performance
  evaluation of face, text, and vehicle detection and tracking in video: Data,
  metrics, and protocol. IEEE Transactions on Pattern Analysis and Machine
  Intelligence  (2008)

\bibitem{kingma2014adam}
Kingma, D.P.: Adam: A method for stochastic optimization. arXiv preprint
  arXiv:1412.6980  (2014)

\bibitem{kuehne2011hmdb}
Kuehne, H., Jhuang, H., Garrote, E., Poggio, T., Serre, T.: Hmdb: a large video
  database for human motion recognition. In: Proceedings of the IEEE/CVF
  International Conference on Computer Vision (2011)

\bibitem{li2021deep}
Li, J., Li, B., Lu, Y.: Deep contextual video compression. In: Advances in
  Neural Information Processing Systems (2021)

\bibitem{li2023neural}
Li, J., Li, B., Lu, Y.: Neural video compression with diverse contexts. In:
  Proceedings of the IEEE/CVF Conference on Computer Vision and Pattern
  Recognition (2023)

\bibitem{li2023uniformer}
Li, K., Wang, Y., Zhang, J., Gao, P., Song, G., Liu, Y., Li, H., Qiao, Y.:
  Uniformer: Unifying convolution and self-attention for visual recognition.
  IEEE Transactions on Pattern Analysis and Machine Intelligence  (2023)

\bibitem{li2022mvitv2}
Li, Y., Wu, C.Y., Fan, H., Mangalam, K., Xiong, B., Malik, J., Feichtenhofer,
  C.: Mvitv2: Improved multiscale vision transformers for classification and
  detection. In: Proceedings of the IEEE/CVF Conference on Computer Vision and
  Pattern Recognition (2022)

\bibitem{li2018resound}
Li, Y., Li, Y., Vasconcelos, N.: Resound: Towards action recognition without
  representation bias. In: European Conference on Computer Vision (2018)

\bibitem{lin2023deepsvc}
Lin, H., Chen, B., Zhang, Z., Lin, J., Wang, X., Zhao, T.: Deepsvc: Deep
  scalable video coding for both machine and human vision. In: ACM
  International Conference on Multimedia (2023)

\bibitem{lin2019tsm}
Lin, J., Gan, C., Han, S.: Tsm: temporal shift module for efficient video
  understanding. In: Proceedings of the IEEE/CVF International Conference on
  Computer Vision (2019)

\bibitem{lin2017feature}
Lin, T.Y., Doll{\'a}r, P., Girshick, R., He, K., Hariharan, B., Belongie, S.:
  Feature pyramid networks for object detection. In: Proceedings of the
  IEEE/CVF Conference on Computer Vision and Pattern Recognition (2017)

\bibitem{liu2020conditional}
Liu, J., Wang, S., Ma, W.C., Shah, M., Hu, R., Dhawan, P., Urtasun, R.:
  Conditional entropy coding for efficient video compression. In: European
  Conference on Computer Vision (2020)

\bibitem{liu2021swin}
Liu, Z., Lin, Y., Cao, Y., Hu, H., Wei, Y., Zhang, Z., Lin, S., Guo, B.: Swin
  transformer: Hierarchical vision transformer using shifted windows. In:
  Proceedings of the IEEE/CVF International Conference on Computer Vision
  (2021)

\bibitem{lu2019dvc}
Lu, G., Ouyang, W., Xu, D., Zhang, X., Cai, C., Gao, Z.: Dvc: An end-to-end
  deep video compression framework. In: Proceedings of the IEEE/CVF Conference
  on Computer Vision and Pattern Recognition (2019)

\bibitem{lu2020end}
Lu, G., Zhang, X., Ouyang, W., Chen, L., Gao, Z., Xu, D.: An end-to-end
  learning framework for video compression. IEEE Transactions on Pattern
  Analysis and Machine Intelligence  (2020)

\bibitem{mentzer2022vct}
Mentzer, F., Toderici, G.D., Minnen, D., Caelles, S., Hwang, S.J., Lucic, M.,
  Agustsson, E.: Vct: A video compression transformer. In: Advances in Neural
  Information Processing Systems (2022)

\bibitem{milan2016mot16}
Milan, A.: Mot16: A benchmark for multi-object tracking. arXiv preprint
  arXiv:1603.00831  (2016)

\bibitem{minnen2020channel}
Minnen, D., Singh, S.: Channel-wise autoregressive entropy models for learned
  image compression. In: IEEE International Conference on Image Processing
  (2020)

\bibitem{oquab2023dinov2}
Oquab, M., Darcet, T., Moutakanni, T., Vo, H., Szafraniec, M., Khalidov, V.,
  Fernandez, P., Haziza, D., Massa, F., El-Nouby, A., et~al.: Dinov2: Learning
  robust visual features without supervision. arXiv preprint arXiv:2304.07193
  (2023)

\bibitem{pan2022st}
Pan, J., Lin, Z., Zhu, X., Shao, J., Li, H.: St-adapter: Parameter-efficient
  image-to-video transfer learning. In: Advances in Neural Information
  Processing Systems (2022)

\bibitem{paszke2019pytorch}
Paszke, A., Gross, S., Massa, F., Lerer, A., Bradbury, J., Chanan, G., Killeen,
  T., Lin, Z., Gimelshein, N., Antiga, L., et~al.: Pytorch: An imperative
  style, high-performance deep learning library. In: Advances in Neural
  Information Processing Systems (2019)

\bibitem{pont20172017}
Pont-Tuset, J., Perazzi, F., Caelles, S., Arbel{\'a}ez, P., Sorkine-Hornung,
  A., Van~Gool, L.: The 2017 davis challenge on video object segmentation.
  arXiv preprint arXiv:1704.00675  (2017)

\bibitem{sandler2018mobilenetv2}
Sandler, M., Howard, A., Zhu, M., Zhmoginov, A., Chen, L.C.: Mobilenetv2:
  Inverted residuals and linear bottlenecks. In: Proceedings of the IEEE/CVF
  Conference on Computer Vision and Pattern Recognition (2018)

\bibitem{shannon1948mathematical}
Shannon, C.E.: A mathematical theory of communication. The Bell system
  technical journal  (1948)

\bibitem{shao2020finegym}
Shao, D., Zhao, Y., Dai, B., Lin, D.: Finegym: A hierarchical video dataset for
  fine-grained action understanding. In: Proceedings of the IEEE/CVF Conference
  on Computer Vision and Pattern Recognition (2020)

\bibitem{singh2020end}
Singh, S., Abu-El-Haija, S., Johnston, N., Ball{\'e}, J., Shrivastava, A.,
  Toderici, G.: End-to-end learning of compressible features. In: IEEE
  International Conference on Image Processing (2020)

\bibitem{soomro2012ucf101}
Soomro, K.: Ucf101: A dataset of 101 human actions classes from videos in the
  wild. arXiv preprint arXiv:1212.0402  (2012)

\bibitem{sullivan2012overview}
Sullivan, G.J., Ohm, J.R., Han, W.J., Wiegand, T.: Overview of the high
  efficiency video coding (hevc) standard. IEEE Transactions on Circuits and
  Systems for Video Technology  (2012)

\bibitem{tan2021diverse}
Tan, Z., Chai, M., Chen, D., Liao, J., Chu, Q., Liu, B., Hua, G., Yu, N.:
  Diverse semantic image synthesis via probability distribution modeling. In:
  Proceedings of the IEEE/CVF Conference on Computer Vision and Pattern
  Recognition (2021)

\bibitem{tian2020self}
Tian, Y., Che, Z., Bao, W., Zhai, G., Gao, Z.: Self-supervised motion
  representation via scattering local motion cues. In: European Conference on
  Computer Vision (2020)

\bibitem{tian2018ban}
Tian, Y., Che, Z., Zhai, G., Gao, Z.: Ban, a barcode accurate detection
  network. In: IEEE Visual Communications and Image Processing (2018)

\bibitem{tian2021self}
Tian, Y., Lu, G., Min, X., Che, Z., Zhai, G., Guo, G., Gao, Z.:
  Self-conditioned probabilistic learning of video rescaling. In: Proceedings
  of the IEEE/CVF International Conference on Computer Vision (2021)

\bibitem{tian2024coding}
Tian, Y., Lu, G., Yan, Y., Zhai, G., Chen, L., Gao, Z.: A coding framework and
  benchmark towards low-bitrate video understanding. IEEE Transactions on
  Pattern Analysis and Machine Intelligence  (2024)

\bibitem{tian2024smc++}
Tian, Y., Lu, G., Zhai, G.: Smc++: Masked learning of unsupervised video
  semantic compression. arXiv preprint arXiv:2406.04765  (2024)

\bibitem{tian2023non}
Tian, Y., Lu, G., Zhai, G., Gao, Z.: Non-semantics suppressed mask learning for
  unsupervised video semantic compression. In: Proceedings of the IEEE/CVF
  International Conference on Computer Vision (2023)

\bibitem{tian2019video}
Tian, Y., Min, X., Zhai, G., Gao, Z.: Video-based early asd detection via
  temporal pyramid networks. In: International Conference on Multimedia and
  Expo (2019)

\bibitem{tian2023clsa}
Tian, Y., Yan, Y., Zhai, G., Chen, L., Gao, Z.: Clsa: a contrastive learning
  framework with selective aggregation for video rescaling. IEEE Transactions
  on Image Processing  (2023)

\bibitem{tian2022ean}
Tian, Y., Yan, Y., Zhai, G., Guo, G., Gao, Z.: Ean: event adaptive network for
  enhanced action recognition. International Journal of Computer Vision  (2022)

\bibitem{tomar2006converting}
Tomar, S.: Converting video formats with ffmpeg. Linux Journal  (2006)

\bibitem{tong2022videomae}
Tong, Z., Song, Y., Wang, J., Wang, L.: Videomae: Masked autoencoders are
  data-efficient learners for self-supervised video pre-training. In: Advances
  in Neural Information Processing Systems (2022)

\bibitem{vaswani2017attention}
Vaswani, A., Shazeer, N., Parmar, N., Uszkoreit, J., Jones, L., Gomez, A.N.,
  Kaiser, {\L}., Polosukhin, I.: Attention is all you need. In: Advances in
  Neural Information Processing Systems (2017)

\bibitem{veselov2021hybrid}
Veselov, A.I., Chen, H., Romano, F., Zhijie, Z., Gilmutdinov, M.R.: Hybrid
  video and feature coding and decoding (2021), uS Patent App. 17/197,500

\bibitem{wang2023look}
Wang, J., Chen, D., Wu, Z., Luo, C., Tang, C., Dai, X., Zhao, Y., Xie, Y.,
  Yuan, L., Jiang, Y.G.: Look before you match: Instance understanding matters
  in video object segmentation. In: Proceedings of the IEEE/CVF Conference on
  Computer Vision and Pattern Recognition (2023)

\bibitem{wang2023videomae}
Wang, L., Huang, B., Zhao, Z., Tong, Z., He, Y., Wang, Y., Wang, Y., Qiao, Y.:
  Videomae v2: Scaling video masked autoencoders with dual masking. In:
  Proceedings of the IEEE/CVF Conference on Computer Vision and Pattern
  Recognition (2023)

\bibitem{wang2021tdn}
Wang, L., Tong, Z., Ji, B., Wu, G.: Tdn: Temporal difference networks for
  efficient action recognition. In: Proceedings of the IEEE/CVF Conference on
  Computer Vision and Pattern Recognition (2021)

\bibitem{wang2022internvideo}
Wang, Y., Li, K., Li, Y., He, Y., Huang, B., Zhao, Z., Zhang, H., Xu, J., Liu,
  Y., Wang, Z., et~al.: Internvideo: General video foundation models via
  generative and discriminative learning. arXiv preprint arXiv:2212.03191
  (2022)

\bibitem{VVenC}
Wieckowski, A., Brandenburg, J., Hinz, T., Bartnik, C., George, V., Hege, G.,
  Helmrich, C., Henkel, A., Lehmann, C., Stoffers, C., Zupancic, I., Bross, B.,
  Marpe, D.: Vvenc: An open and optimized vvc encoder implementation. In: IEEE
  International Conference on Multimedia and Expo Workshops

\bibitem{wiegand2003overview}
Wiegand, T., Sullivan, G.J., Bjontegaard, G., Luthra, A.: Overview of the h.
  264/avc video coding standard. IEEE Transactions on Circuits and Systems for
  Video Technology  (2003)

\bibitem{wu2018video}
Wu, C.Y., Singhal, N., Krahenbuhl, P.: Video compression through image
  interpolation. In: European Conference on Computer Vision (2018)

\bibitem{xie2021self}
Xie, Z., Lin, Y., Yao, Z., Zhang, Z., Dai, Q., Cao, Y., Hu, H.: Self-supervised
  learning with swin transformers. arXiv preprint arXiv:2105.04553  (2021)

\bibitem{xu2015empirical}
Xu, B., Wang, N., Chen, T., Li, M.: Empirical evaluation of rectified
  activations in convolutional network (2015). arXiv preprint arXiv:1505.00853
  (2015)

\bibitem{yan2021dehib}
Yan, Z., Li, G., Tian, Y., Wu, J., Li, S., Chen, M., Poor, H.V.: Dehib: Deep
  hidden backdoor attack on semi-supervised learning via adversarial
  perturbation. In: Annual AAAI Conference on Artificial Intelligence (2021)

\bibitem{dhbe_asiaccs_2023}
Yan, Z., Li, S., Zhao, R., Tian, Y., Zhao, Y.: Dhbe: Data-free holistic
  backdoor erasing in deep neural networks via restricted adversarial
  distillation. In: ACM ASIA Conference on Computer and Communications Security
  (2023)

\bibitem{yang2020variable}
Yang, F., Herranz, L., Van De~Weijer, J., Guiti{\'a}n, J.A.I., L{\'o}pez, A.M.,
  Mozerov, M.G.: Variable rate deep image compression with modulated
  autoencoder. IEEE Signal Processing Letters  (2020)

\bibitem{yang2020learning}
Yang, R., Mentzer, F., Van~Gool, L., Timofte, R.: Learning for video
  compression with recurrent auto-encoder and recurrent probability model. IEEE
  Journal of Selected Topics in Signal Processing  (2020)

\bibitem{yang2022advancing}
Yang, R., Timofte, R., Van~Gool, L.: Advancing learned video compression with
  in-loop frame prediction. IEEE Transactions on Circuits and Systems for Video
  Technology  (2022)

\bibitem{yang2021perceptual}
Yang, R., Timofte, R., Van~Gool, L.: Perceptual learned video compression with
  recurrent conditional gan. In: IJCAI (2022)

\bibitem{yang2020discernible}
Yang, Z., Wang, Y., Xu, C., Du, P., Xu, C., Xu, C., Tian, Q.: Discernible image
  compression. In: ACM International Conference on Multimedia (2020)

\bibitem{yi2021benchmarking}
Yi, C., Yang, S., Li, H., Tan, Y.p., Kot, A.: Benchmarking the robustness of
  spatial-temporal models against corruptions. In: Advances in Neural
  Information Processing Systems (2021)

\bibitem{yi2021attention}
Yi, F., Chen, M., Sun, W., Min, X., Tian, Y., Zhai, G.: Attention based network
  for no-reference ugc video quality assessment. In: IEEE International
  Conference on Image Processing (2021)

\bibitem{zhang2011parametric}
Zhang, F., Bull, D.R.: A parametric framework for video compression using
  region-based texture models. IEEE Journal of Selected Topics in Signal
  Processing  (2011)

\bibitem{zhang2021just}
Zhang, Q., Wang, S., Zhang, X., Ma, S., Gao, W.: Just recognizable distortion
  for machine vision oriented image and video coding. International Journal of
  Computer Vision  (2021)

\bibitem{zhang2018unreasonable}
Zhang, R., Isola, P., Efros, A.A., Shechtman, E., Wang, O.: The unreasonable
  effectiveness of deep features as a perceptual metric. In: Proceedings of the
  IEEE/CVF Conference on Computer Vision and Pattern Recognition (2018)

\bibitem{zhang2023lvqac}
Zhang, X., Wu, X.: Lvqac: Lattice vector quantization coupled with spatially
  adaptive companding for efficient learned image compression. In: Proceedings
  of the IEEE/CVF Conference on Computer Vision and Pattern Recognition (2023)

\bibitem{zhang2016joint}
Zhang, X., Ma, S., Wang, S., Zhang, X., Sun, H., Gao, W.: A joint compression
  scheme of video feature descriptors and visual content. IEEE Transactions on
  Image Processing  (2016)

\bibitem{zhang2024gaussianimage}
Zhang, X., Ge, X., Xu, T., He, D., Wang, Y., Qin, H., Lu, G., Geng, J., Zhang,
  J.: Gaussianimage: 1000 fps image representation and compression by 2d
  gaussian splatting. arXiv preprint arXiv:2403.08551  (2024)

\bibitem{zhang2021bytetrack}
Zhang, Y., Sun, P., Jiang, Y., Yu, D., Weng, F., Yuan, Z., Luo, P., Liu, W.,
  Wang, X.: Bytetrack: Multi-object tracking by associating every detection
  box. In: European Conference on Computer Vision (2022)

\bibitem{zhaovideoprism}
Zhao, L., Gundavarapu, N.B., Yuan, L., Zhou, H., Yan, S., Sun, J.J., Friedman,
  L., Qian, R., Weyand, T., Zhao, Y., et~al.: Videoprism: A foundational visual
  encoder for video understanding. In: International Conference on Machine
  Learning (2024)

\bibitem{zhou2021ibot}
Zhou, J., Wei, C., Wang, H., Shen, W., Xie, C., Yuille, A., Kong, T.: ibot:
  Image bert pre-training with online tokenizer. In: International Conference
  on Learning Representations (2021)

\end{thebibliography}
\end{document}